\documentclass[lettersize,journal]{IEEEtran}
\usepackage{amsmath,amsfonts}
\usepackage{algorithmic}
\usepackage{algorithm}
\usepackage{array}

\usepackage[caption=false,font=footnotesize,labelfont=rm,textfont=rm]{subfig}
\usepackage{textcomp}
\usepackage{stfloats}
\usepackage{url}
\usepackage{verbatim}
\usepackage{graphicx}
\usepackage{cite}

\usepackage{multirow}
\usepackage{multicol}
\usepackage{colortbl}
\usepackage{bm}
\usepackage{soul}
\usepackage{color, xcolor}
\usepackage{url}
\usepackage[colorlinks]{hyperref}

\begin{document}

\title{Bi-C\textsuperscript{2}R: Bidirectional Continual Compatible Representation for Re-indexing Free Lifelong Person Re-identification}

\author{
\IEEEauthorblockN{Zhenyu~Cui, Jiahuan~Zhou, \IEEEmembership{Member,~IEEE}, and Yuxin~Peng, \IEEEmembership{Fellow,~IEEE}}

\thanks{
This work was supported by the grants from the National Natural Science Foundation of China (62525201, 62132001, 62432001) and Beijing Natural Science Foundation (L247006).

Zhenyu Cui, Jiahuan Zhou and Yuxin Peng are with the Wangxuan Institute of Computer Technology, Peking University, Beijing 100871, China.

Corresponding author: Yuxin Peng (e-mail: pengyuxin@pku.edu.cn).
}
}

\markboth{IEEE Transactions on Pattern Analysis and Machine Intelligence}%
{Shell \MakeLowercase{\textit{Cui et al.}}: Bi-C\textsuperscript{2}R: Bidirectional Continual Compatible Representation for Re-indexing Free Lifelong Person Re-identification}

\maketitle

\begin{abstract}
Lifelong person Re-IDentification (L-ReID) exploits sequentially collected data to continuously train and update a ReID model, focusing on the overall performance of all data. Its main challenge is to avoid the catastrophic forgetting problem of old knowledge while training on new data. Existing L-ReID methods typically re-extract new features for all historical gallery images for inference after each update, known as ``re-indexing''. However, historical gallery data typically suffers from direct saving due to the data privacy issue and the high re-indexing costs for large-scale gallery images. As a result, it inevitably leads to incompatible retrieval between query features extracted by the updated model and gallery features extracted by those before the update, greatly impairing the re-identification performance. To tackle the above issue, this paper focuses on a new task called Re-index Free Lifelong person Re-IDentification (RFL-ReID), which requires performing lifelong person re-identification without re-indexing historical gallery images. Therefore, RFL-ReID is more challenging than L-ReID, requiring continuous learning and balancing new and old knowledge in diverse streaming data, and making the features output by the new and old models compatible with each other. To this end, we propose a Bidirectional Continuous Compatible Representation (Bi-C\textsuperscript{2}R) framework to continuously update the gallery features extracted by the old model to perform efficient L-ReID in a compatible manner. Specifically, a bidirectional compatible transfer network is first designed to bridge the relationship between new and old knowledge and continuously update the old gallery features to the new feature space after the updating. Secondly, a bidirectional compatible distillation module and a bidirectional anti-forgetting distillation model are designed to balance the compatibility between the new and old knowledge in dual feature spaces. Finally, a feature-level exponential moving average strategy is designed to adaptively fill the diverse knowledge gaps between different data domains. Finally, we verify our proposed Bi-C\textsuperscript{2}R method through theoretical analysis and extensive experiments on multiple benchmarks, which demonstrate that the proposed method can achieve leading performance on both the introduced RFL-ReID task and the traditional L-ReID task. The source code of this paper is available at \href{https://github.com/PKU-ICST-MIPL/Bi-C2R-TPAMI2026}{https://github.com/PKU-ICST-MIPL/Bi-C2R-TPAMI2026}.
\end{abstract}


\begin{figure}[t!]
\begin{center}
\includegraphics[width=0.97\linewidth]{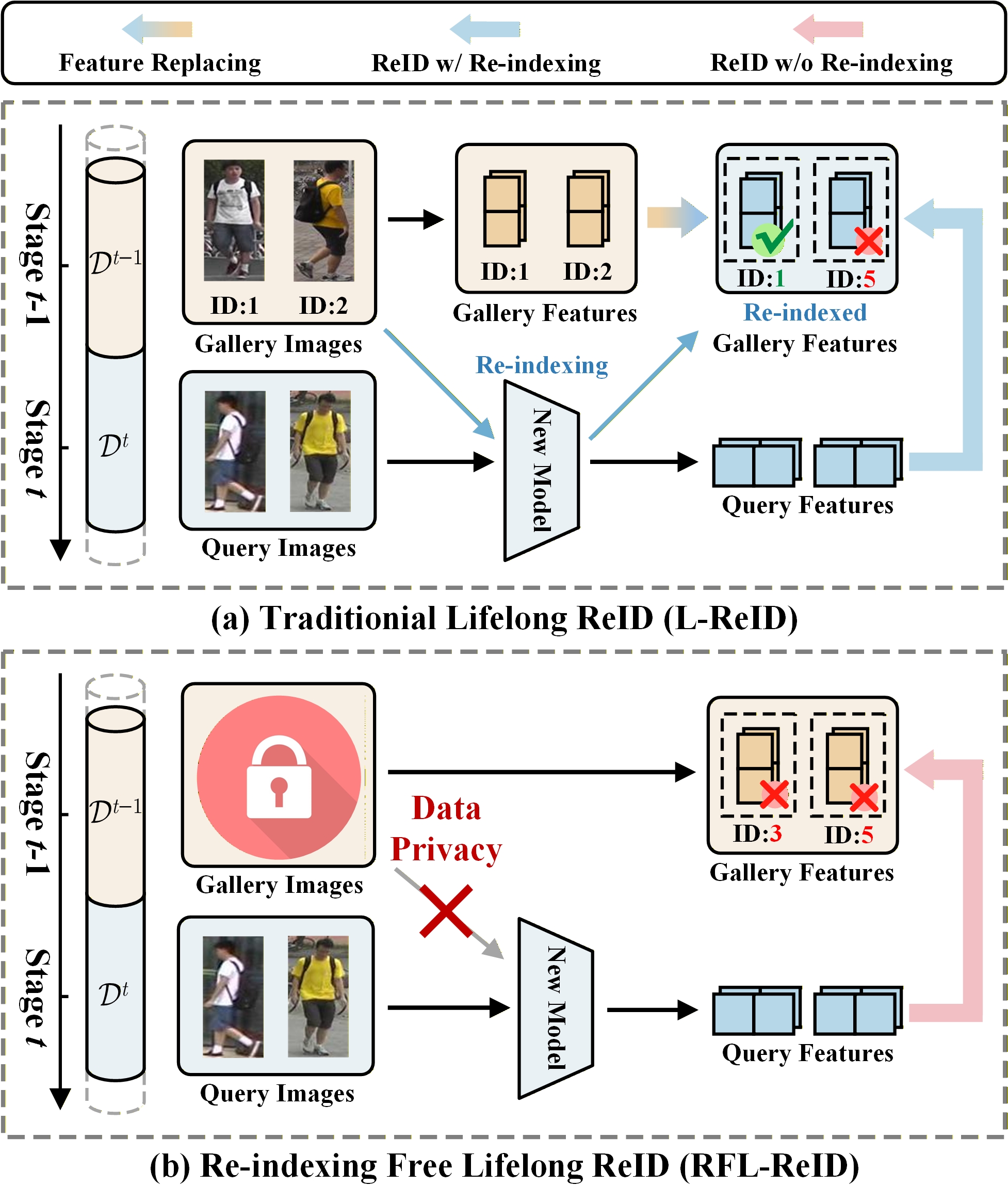}
\end{center}
\vspace{-10pt}
\caption{Comparison between the Traditional Lifelong person Re-identification (L-ReID) task (a) and our introduced Re-indexing Free Lifelong person Re-identification (RFL-ReID) task (b).}
\vspace{-5pt}
\label{fig: task}
\end{figure}

\begin{figure}[t!]
\begin{center}
\includegraphics[width=0.97\linewidth]{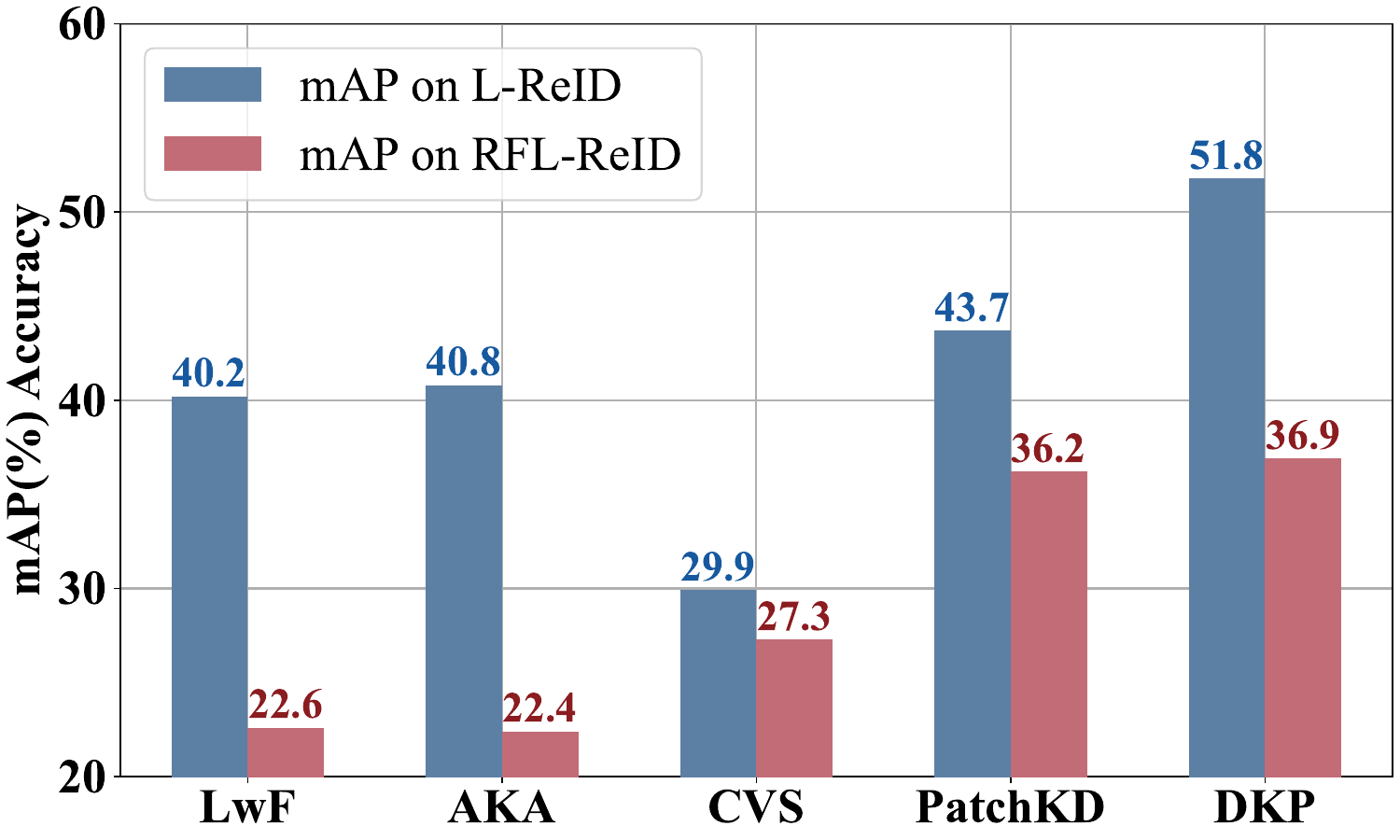}
\end{center}
\vspace{-10pt}
\caption{Performance Comparison of existing methods, including LwF~\cite{li2017learning}, AKA~\cite{pu2021lifelong}, CVS~\cite{wan2022continual}, PatchKD~\cite{sun2022patch}, DKP~\cite{xu2024distribution}.}
\vspace{-5pt}
\label{fig: perform_decay}
\end{figure}

\section{Introduction}
\par \IEEEPARstart{A}{s} a ‌fundamental‌ vision task, person Re-IDentification (ReID) pursues to identify the same individual across non-overlapping camera views over different periods~\cite{zheng2016person}. Benefiting from pre-collected large-scale datasets~\cite{market1501,dukemtmc,msmt17v2} recording massive people with identity annotations, many advancements~\cite{wang2016joint, zhong2017re, sun2018beyond, ye2021deep} have achieved remarkable performance and demonstrated promising progress in multiple scenarios, \emph{e.g.}, Cloth-Changing ReID~\cite{li2024disentangling, nguyen2024temporal}, Visible-Infrared ReID~\cite{ren2024implicit, yang2024shallow}, etc. However, real data from different scenarios are typically collected continuously, demanding that the ReID model continuously adapt to new data. As a result, the performance of existing methods usually degrades severely due to the well-known catastrophic forgetting problem~\cite{de2021continual}.

\begin{figure}[t!]
\begin{center}
\includegraphics[width=0.97\linewidth]{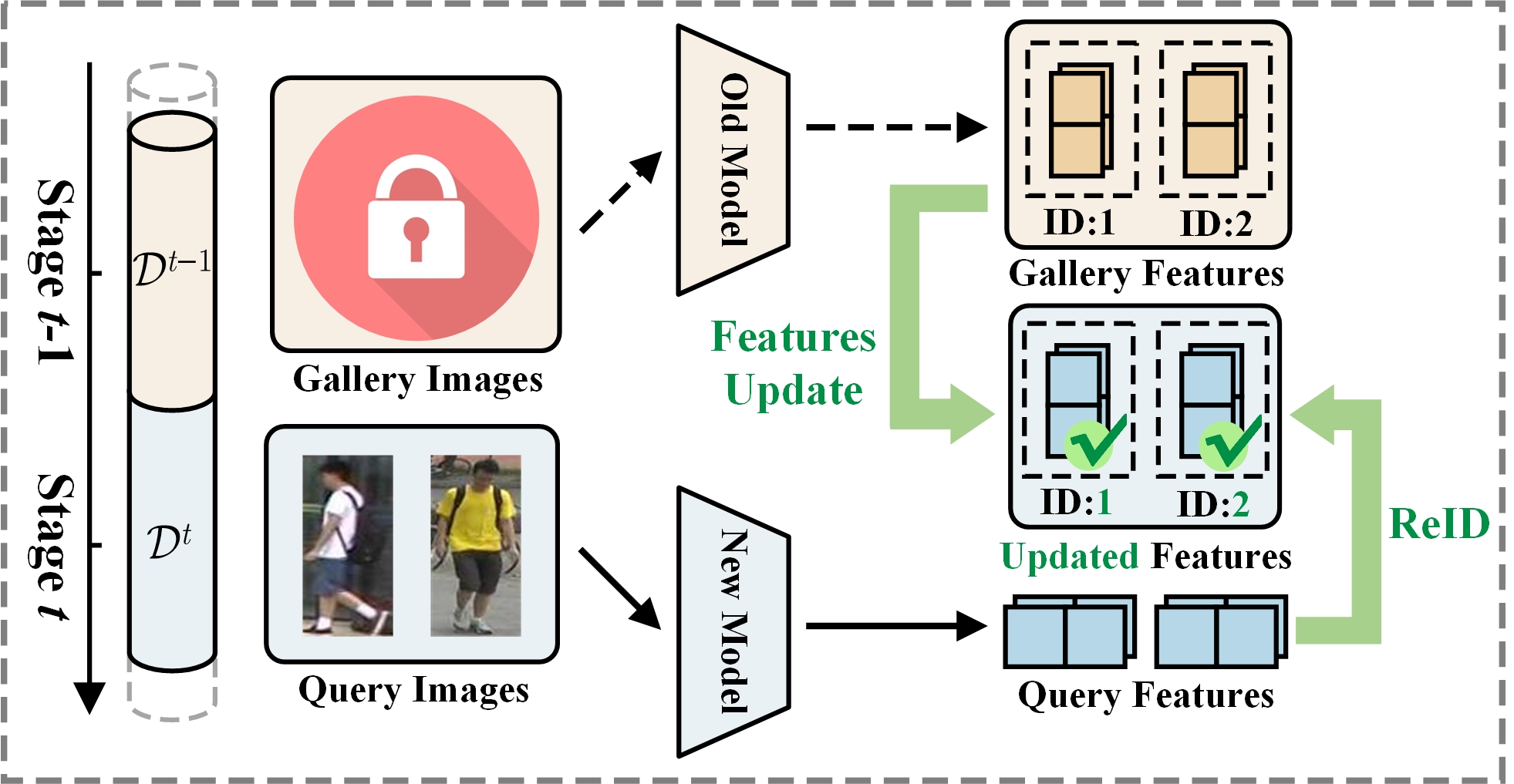}
\end{center}
\caption{A schematic diagram of our proposed Bidirectional Continual Compatible Representation (Bi-C\textsuperscript{2}R) method.}
\label{fig: motivation}
\end{figure}

\par To this end, Lifelong person Re-IDentification (L-ReID) has emerged as a promising research hotspots~\cite{pu2021lifelong, yu2023lifelong, ge2022lifelong, pu2022meta}, focusing on sequentially learning from streaming data while ensuring robust overall performance across both new and historical data. The core challenge of L-ReID lies in achieving an efficient balance between old and new knowledge. Existing L-ReID methods can be predominantly categorized into exemplar-based methods~\cite{yu2023lifelong, ge2022lifelong} and non-exemplar-based methods~\cite{sun2022patch, xu2024lstkc, xu2024distribution}, which updating the ReID model with knowledge distillation techniques~\cite{gou2021knowledge} to mitigate the catastrophic forgetting problem of old knowledge. Sequentially, the updated model is usually employed to re-extract features of the historical gallery images, which makes it consistent with the query image features, also called ``re-indexing''~\cite{shen2020towards}.
Specifically, Fig.~\ref{fig: task} (a) demonstrates the re-indexing process where the updated ReID model can benefit from the mutually compatible query and gallery features extracted by the same model. Unfortunately, it is impractical to store and re-indexing historical gallery images in a real large-scale L-ReID system due to the huge cost of original image storage and the unbearable computational cost of the updating~\cite{zhang2022towards,ramanujan2022forward}. In particular, considering the privacy of individual data, re-indexing is legally forbidden due to the potential unauthorized storing of original images in the privacy-sensitive scenario~\cite{meng2021learning,ahmad2022event,erkin2009privacy}. To this end, as shown in Fig.~\ref{fig: task} (b), this paper focuses on a practical but more challenging task that prevents the L-ReID model from re-indexing historical gallery images, also called Re-indexing Free Lifelong ReID (RFL-ReID).

\par Compared with L-ReID, RFL-ReID is more challenging when it comes to the large domain gap between continuously collected streaming data, where the difference between new and old knowledge makes the historical gallery features suffer from incompatibility with the updated query features when balancing differential knowledge. Consequently, as shown in Fig.~\ref{fig: perform_decay}, such incompatibility usually leads to serious degradation of existing methods in the RFL-ReID task. To tackle the above issue, compatible training (CT)~\cite{ramanujan2022forward,seo2023metric,pan2023boundary} becomes a primary solution that enables the features extracted by ReID models trained on dissimilar data to be compatible with each other. Unfortunately, existing CT methods failed to apply to the RFL-ReID task in two aspects. First, it demands compatibility across multiple continuously updated ReID models, while existing CT methods only focus on the compatibility between two models~\cite{seo2023metric}, which limits their flexibility in lifelong scenarios. Besides, existing CT methods only focus on compatibility within the same dataset~\cite{pan2023boundary}, ignoring the domain shift between continuously collected variant data.

\par To migrate the aforementioned issues, this paper focuses on a practical but challenging task called Re-indexing Free Lifelong person Re-IDentification (RFL-ReID), involving updating ReID models without re-indexing historical gallery images. Beyond traditional L-ReID, RFL-ReID emphasizes updating the ReID model using streaming data, as well as enabling their compatibility with each other. Therefore, continuously updating the historical gallery features through forward transfer becomes a feasible solution. However, considering the complementarity and discrepancy of different domains, it is challenging to achieve long-term compatibility and anti-forgetting. Specifically, considering the complementarity between new and old domain knowledge, the forward transfer from old features to new feature space ignores the purification of new discriminative information to historical features, limiting the accumulation of diverse knowledge. In addition, the diverse discrepancy limits the forward transfer to comprehensively retain the long-term compatibility of the historical gallery features, leading to mediocre RFL-ReID overall performance.

\par To this end, we propose a Bidirectional Continual Compatible Representation (Bi-C\textsuperscript{2}R) framework for RFL-ReID. As illustrated by Fig.~\ref{fig: motivation}, our Bi-C\textsuperscript{2}R aims to continuously update historical gallery features to alleviate the incompatibility between updated query features and gallery features without re-indexing. To tackle the incompatibility brought by the domain shift problem, we propose a Bidirectional Compatible Transfer Network (BiCT-Net) to transfer old gallery features into new feature space continuously while reconstructing old features based on the new features to prevent the catastrophic forgetting problem from overfitting to new data. Correspondingly, to promote its capacity to effectively balance varied knowledge, a Bidirectional Compatible Distillation (BiCD) module and a Bidirectional Anti-forgetting Distillation (BiAD) module are designed to achieve compatibility between the transferred features with the new features and its capacity to preserve the old knowledge. Furthermore, considering the varied discrepancy between different domains, we propose a Dynamic Feature Fusion (DFF) module to adaptively balance the transferred and old features through knowledge changes, avoiding the dynamic forgetting of old knowledge derived from variant updating difficulty. The main contributions of this paper are four-folded:

\begin{itemize}
\item We propose a Bidirectional Continual Compatible Representation (Bi-C\textsuperscript{2}R) framework to solve a practical but challenging task named Re-indexing Free Lifelong Person Re-identification (RFL-ReID), which requires performing lifelong person re-identification without re-indexing historical gallery images.
\item A Bidirectional Compatible Transfer network (BiCT-Net) is designed to achieve compatibility between new and old features through consistent forward and backward transfer. Correspondingly, a Bidirectional Compatible Distillation (BiCD) module and a Bidirectional Anti-forgetting Distillation (BiAD) module are designed to achieve the compatibility through alignment of both feature distributions, while balancing the new and old knowledge through the distillation of dual discriminative knowledge consolidation.
\item To achieve a balance between varied discrepancies between different domains, a Dynamic Feature Fusion (DFF) module is designed to fuse the old features and the transferred features adaptively, effectively alleviating the cumulative error of forgetting old knowledge derived from continuously compatible transfer.
\item In-depth theoretical analyses and extensive experiments demonstrate the superiority of our proposed Bi-C\textsuperscript{2}R in old knowledge anti-forgetting, new and old features compatibility, and different domain adaptation over seven datasets on nine orders, including Market1501~\cite{market1501}, CUHK-SYSU~\cite{cuhksysu}, DukeMTMC~\cite{dukemtmc}, MSMT17~\cite{msmt17v2}, CUHK03~\cite{cuhk03}, LTCC~\cite{qian2020long}, and PRCC~\cite{yang2019person}.
\end{itemize}

\par This manuscript is an extension of our previous conference paper~\cite{cui2024learning}, while we have made plenty of extensions including 1) To facilitate the new knowledge from updating old features and the capturing of new knowledge, we extend the forward feature transfer network into a bidirectional compatible transfer network and two bidirectional distillation modules are explored to bridge both new and old knowledge from mutual knowledge transfer. 2) To further improve the adaptability to variant updating difficulty, we propose a dynamic feature fusion module to eliminate the cumulative error derived from long-term knowledge forgetting. 3) Thorough theoretical analyses of our proposed bidirectional transfer process and the dynamic fusion module demonstrate the superiority of our method against existing methods. 4) Experiments on more training orders compared with more state-of-the-art methods on nine L-ReID benchmarks consists of seven popular person re-identification datasets are present to verify the effectiveness of our method. 5) More ablation study and visualization results intuitively illustrated the effectiveness of each component of our proposed Bi-C\textsuperscript{2}R.

\section{Related Work}
\subsection{Person Re-identification}
\par Person Re-IDentification (ReID) aims to match the same person in different locations with their visual information captured by different cameras~\cite{ye2021deep, zheng2016person, fu2022domain, tan2017person}. Its core challenge involves extracting unique discriminative features to match individuals and avoiding suffering from lighting~\cite{huang2019illumination}, posture~\cite{gu2020appearance}, background~\cite{tian2018eliminating}, etc. To overcome the increasingly complex ReID scenarios, recent advancements have focused on more challenging ReID tasks, \emph{e.g.}, Chloth-changing ReID~\cite{qian2020long,gu2022clothes,cui2023dcr}, Visible-infrared ReID~\cite{wang2019learning, cui2024dma}, Video-based ReID~\cite{mclaughlin2016recurrent, yang2020spatial}, etc., committing to matching the same pedestrian across intricate periods and locations. Despite some progress, these methods usually require the pre-collection of ‌‌massive individual images with extensive annotations for training, which makes it difficult to build a ReID model that can handle data from different scenarios.

\subsection{Lifelong Person Re-identification}

\par As one of the classic incremental learning task~\cite{zhang2024fscil}, Lifelong person Re-IDentification (L-ReID) aims to continuously train a ReID model using sequentially collected data from different scenarios, achieving person matching across different scenarios~\cite{pu2021lifelong}. The challenge of L-ReID lies in learning new knowledge in new scenarios while preserving the old ones to mitigate the catastrophic forgetting problem. To tackle this challenge, existing L-ReID methods have made progress on two aspects: exemplar-based L-ReID methods~\cite{wu2021generalising,yu2023lifelong,ge2022lifelong} and non-exemplar L-ReID methods~\cite{pu2021lifelong,pu2022meta,sun2022patch,wang2022positive,xu2024lstkc,xu2024distribution}.

\par The exemplar-based L-ReID methods maintain a memory bank for old data when learning new data, and achieve anti-forgetting of old knowledge through joint learning of both new and memorized old data. Considering the huge domain gap within streaming data, Ge et al.~\cite{ge2022lifelong} proposed a pseudo task knowledge preservation method, which reconstructs old features with replay data to reduce the discrepancy between new and old knowledge. It also explores a domain consistency loss to effectively preserve task-shared knowledge by jointly distilling discriminative identity information. To further incorporate the discriminative information from new and old knowledge, Yu et al.~\cite{yu2023lifelong} designed a knowledge refreshing and consolidation pipeline to allow both new and old knowledge. It exploits two models to learn new and old knowledge respectively and guides the fusion of both models by the mutual transferring of new and old knowledge, which promotes the discriminability of both new and old knowledge to perform well in all data. Despite some progress, their performance is typically severely degraded without old training data rehearsing due to data privacy issues, over-dependent on old data and lack of necessary anti-forgetting measures.

\par To address the above challenges, exemplar-based L-ReID methods are devoted to alleviating the catastrophic forgetting problem without rehearsing old training data~\cite{pu2021lifelong,pu2022meta,sun2022patch,wang2022positive,xu2024lstkc,xu2024distribution}. To this end, Pu et al.~\cite{pu2021lifelong} designed an adaptive knowledge accumulation framework for knowledge representation and operation, which employs the previous knowledge to the current stage by representing transferable knowledge as a knowledge graph, avoiding the time-consuming replay of old data. To preserve fine-grained discriminative information, Sun et al.~\cite{sun2022patch} introduced a patch-based knowledge distillation method. It performs effective knowledge preservation by distilling patch-level local information, thus mitigating the background clutters under the distribution shift. Considering the gap in learning knowledge between different domains, Xu et al.~\cite{xu2024lstkc} proposed a long short-term knowledge consolidation method. It facilitates short-term knowledge anti-forgetting by filtering out erroneous old knowledge when performing distillation. Meanwhile, an adaptive parameter fusion method is designed to preserve long-term knowledge. Consequentially, they further improve the new knowledge capture capacity by identity distribution modelling~\cite{xu2024lstkc}, which achieves accurate instance-level representation for each identity by modelling each person as an anisotropic feature distribution, where identity-level can be continuously refined and achieving remarkable performance.

\par However, although the state-of-the-art L-ReID methods avoid replaying training data when performing updating, they still violate data privacy due to the requirement of re-extracting the data in the old gallery. Therefore, in this paper, we focus on such a challenging task called RFL-ReID, which avoids data privacy issues derived from repeatedly extracting features of data in the old gallery.

\begin{figure*}[ht]
\begin{center}
\includegraphics[width=\linewidth]{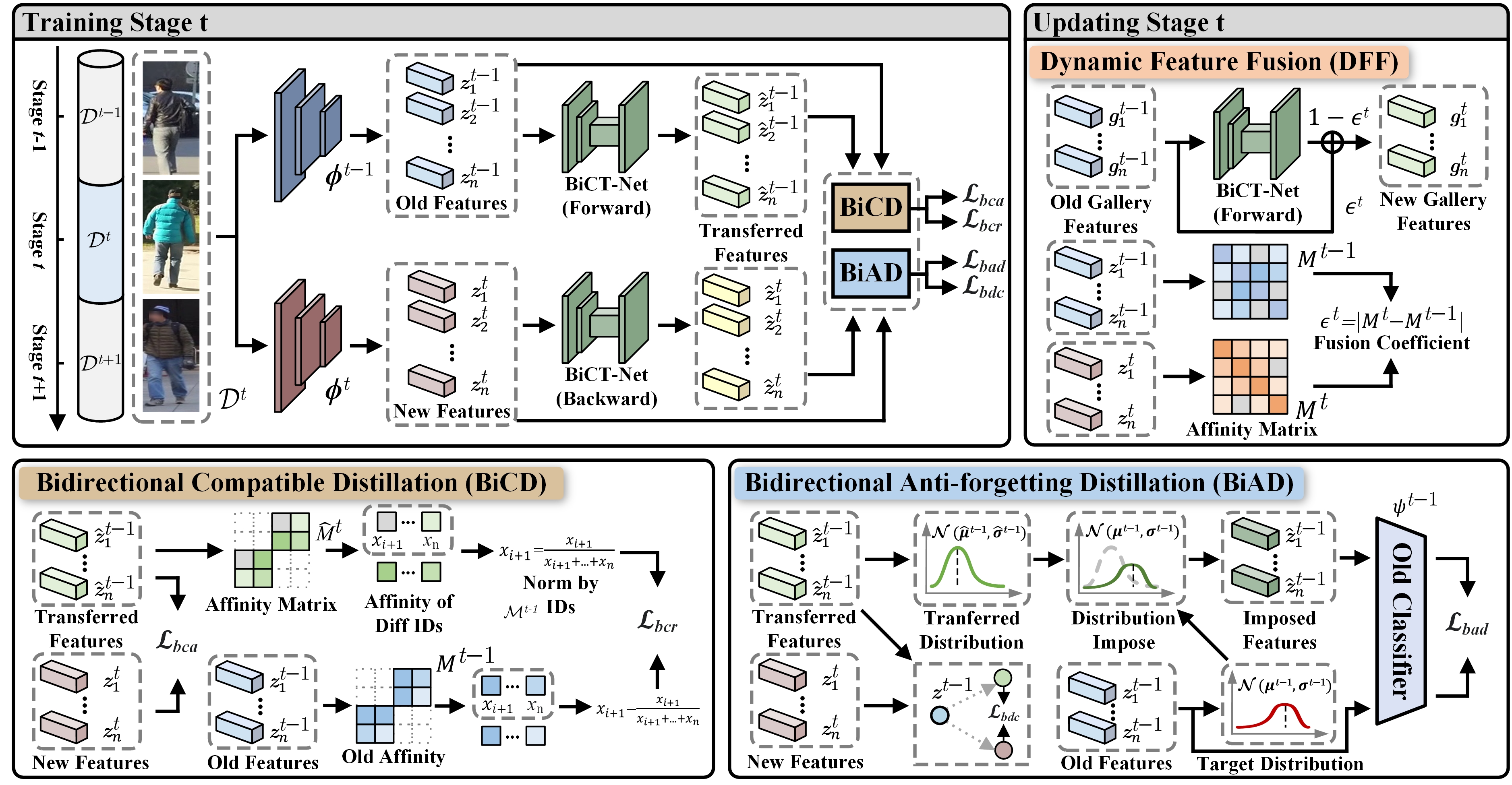}
\end{center}
\caption{The architecture of our proposed Bidirectional Continual Compatible Representation (Bi-C\textsuperscript{2}R) method. Our Bi-C\textsuperscript{2}R consists of a Bidirectional (Forward/Backward) Compatible Transfer network (BiCT-Net), a Bidirectional Compatible Distillation (BiCD) module, and a Bidirectional Anti-forgetting Distillation (BiAD) module. In the training phase, all the above components will be trained, and the BiCT-Net (forward) will be used to update the old feature set. In the updating phase, a Dynamic Feature Fusion (DFF) module is employed to update the old gallery features.}
\label{fig: method}
\end{figure*}

\subsection{Compatible Training}
\par Compatible Training (CT) aims to enable features extracted by two different models to be mutually retrieved without re-indexing data in the old gallery. Existing CT methods focus on updating the model using homologous data and achieving compatibility by reducing the discrepancy between the features extracted by the two models before and after the update. 

\par To this end, backward CT methods~\cite{shen2020towards,zhang2022towards,pan2023boundary,liang2023mixbct} aim to constrain the features extracted by the updated model to the feature space generated by the model before the update. Zhang et al.~\cite{zhang2022towards} proposed a unified compatible representation framework, which improved the old class centroids by structural prototype refinement and tackled the challenging data splits problem. Pan et al.~\cite{pan2023boundary} proposed to reduce the distribution disparity between new and old models through adversarial optimization and improve the compatibility and discriminability of the model through boundary constraints. However, due to the excessive restriction of the old feature space, backward CT methods typically sacrifice the plasticity of the updated model.

\par Therefore, forward CT methods~\cite{zhou2022forward,ramanujan2022forward,wan2022continual} align the features extracted by the old model with the new one through appropriate transformations, thereby improving the performance on new data. Ramanujan et al.~\cite{ramanujan2022forward} proposed to use side-information for forward compatibility, which greatly improves its flexibility. However, the above methods ignore the requirement for compatibility between continuously collected datasets, exacerbating its catastrophic forgetting problem. To this end, Wan et al.~\cite{wan2022continual} proposed a continual visual search framework for compatibility in lifelong scenarios, allowing new classes to appear in the continuously collected data. However, it severely relies on rehearsing historical data and models, which is strictly prohibited under the data privacy conditions.

\par Different from these methods, we propose a bidirectional continual compatible representation framework, called Bi-C\textsuperscript{2}R. It achieves advanced L-ReID performance when old data in the gallery cannot be re-indexed in the practical privacy-sensitive scenario.

\begin{figure}[htbp!]
\begin{center}
\includegraphics[width=\linewidth]{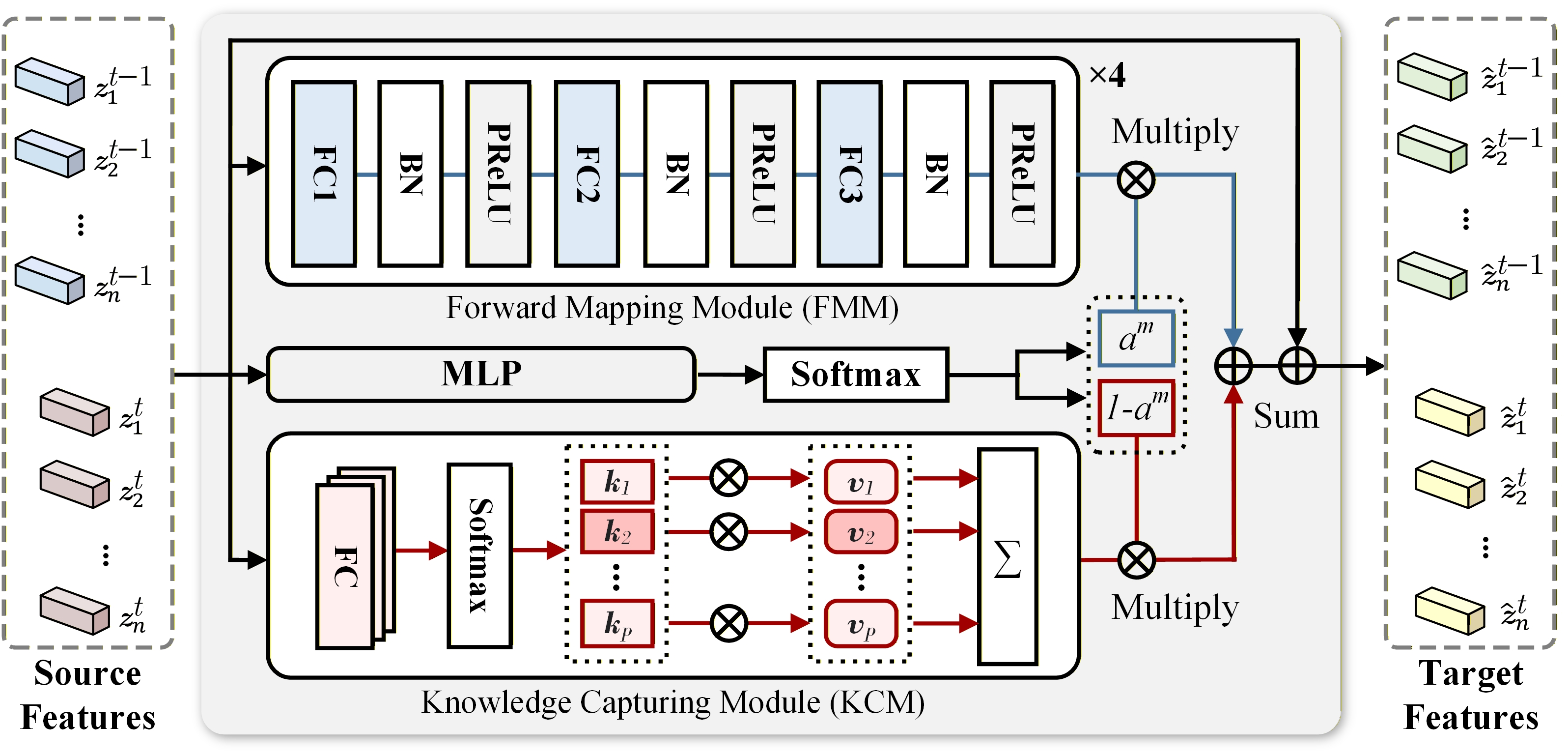}
\end{center}
\caption{The structure of our proposed Bidirectional Compatible Transfer network (BiCT-Net).}
\label{fig: bict}
\end{figure}

\section{The Proposed Method}

\subsection{Problem Settings and Notations}
\par We start by formally defining our introduced Re-indexing Free Lifelong person Re-identification (RFL-ReID) task. Given a sequence of datasets $\mathcal{D}=\{\mathcal{D}^{1}, \mathcal{D}^{2}, ..., \mathcal{D}^{T}\}$ sequentially collected continuously in $T$ stages, the $t^{th}$ dataset $\mathcal{D}^{t}=\{x_{i}^{t},y_{i}^{t}\}^{N^t}_{i=1}$ contains $N^t$ training images $x_{i}^{t}$ and the corresponding identity labels $y_{i}^{t}$. For convenience, each $\mathcal{D}^{t}$ is randomly divided into a training set $\mathcal{T}^{t}$ and a gallery set $\mathcal{G}^{t}$ without overlapped identities, while a query set $\mathcal{Q}^{t}$ is also provided for testing. Derived from L-ReID, RFL-ReID requires to use the streaming datasets $(\mathcal{T}^{1}, ..., \mathcal{T}^{T})$ to sequentially train a ReID model $\phi^T(\cdot)$ across $T$ stages, which prohibits the availability of the previous $t-1$ datasets $(\mathcal{T}^{1}, ..., \mathcal{T}^{t-1})$ when training on $\mathcal{T}^{t}$ at the $t^{th}$ stage. For inference, the trained model $\phi^t(\cdot)$ is expected to maximize the features of the same person from every paired gallery and query set $\{(\mathcal{G}^{i}, \mathcal{Q}^{i})\}^{t}_{i=1}$ at the $t^{th}$ stage. Notably, due to data privacy issues, images of the previous $(t-1)^{th}$ stages are not allowed to be stored in $\mathcal{G}^{t}$, where the gallery features of $\mathcal{G}^{t-1}$ can only be extracted by $\phi^{t-1}(\cdot)$ trained after the $t-1^{th}$ stage, while the query features are extracted by $\phi^t(\cdot)$ after all $t$ stages. Finally, the similarity between each query feature and all gallery features across all $t$ stages forms a rank list, which is employed to verify the ReID performance.

\subsection{Overview of the Proposed Method}
\par Fig.~\ref{fig: method} illustrates our proposed Bidirectional Continual Compatible Representation (Bi-C\textsuperscript{2}R) method, which consists of two Bidirectional (Forward/Backward) Continual Compatible Transfer Networks (BiCT-Net), a Bidirectional Compatible Distillation (BiCD) module, a Bidirectional Anti-forgetting Distillation (BiAD) module, and a Dynamic Feature Fusion (DFF) module. When starting training at the $t^{th}$ training stage, we froze the latest feature extraction model $\phi^{t-1}$ with the latest classifier $\psi^{t-1}$ as the old knowledge. Considering that the fundamental purpose of the RFL-ReID task is to capture new knowledge while avoiding the catastrophic forgetting of old knowledge, a baseline method~\cite{xu2024lstkc} is employed to train a current model $\phi^t$ initialized with $\psi^{t-1}$ to form a basic discriminative representation for each person. To enable the old model $\phi^{t-1}$ to be compatible with the current one, we further train the BiCT-Net using the BiCD and the BiAD modules at the training phase to further balance the compatibility between the old and new knowledge. In the updating phase, we construct a new gallery feature set $\mathcal{G}^{t}$ by updating the historical gallery feature set $\mathcal{G}^{t-1}$ using the forward BiCT-Net with the DFF module, and extracting the features of the current gallery set $\mathcal{G}^{t}$ using $\phi^t$. Therefore, when a query image arrives, the ranking of all historical gallery images can be directly calculated without repeating the re-indexing process.

\par Notably, in our previous conference paper~\cite{cui2024learning}, we proposed the forward transfer from old features to new feature space as a preliminary method C\textsuperscript{2}R, which consists of the forward transfer versions of BiCT-Net, BiCD and BiAD, i.e., CCT network, BCD module, and BAD module, respectively. In the following sections, we detail the above modules in the following sections. For convenience, we detail the above modules in the following section from the perspective of forward transfer, which performs the same process for backward transfer.

\subsection{Bidirectional Compatible Transfer network}

\par To tackle the domain shift problem, we design a Bidirectional Compatible Transfer network (BiCT-Net) to perform mutual transfer between the target feature and the source feature continuously. To this end, as shown in Fig.~\ref{fig: bict}, our BiCT network is designed as a two-stream network to capture transfer-related knowledge from different domains.

\par Specifically, we first design a Knowledge Capturing Module (KCM) module to capture knowledge from the target domain. Let $\bm{z}^{t-1} \in \mathbb{R}^{C}$ be the feature output by the old model $\phi^{t-1}(\cdot)$ based on the sample $x^{t-1}$, $\bm{v}_p\in \mathbb{R}^{P \times C}$ be a learnable new knowledge prototype set with length $P$. In order to enable the output feature $\bm{\widehat{z}}^{t-1}$ to capture new knowledge for compatibility, we first calculate a capturing probability $\bm{k}\in \mathbb{R}^{P}$ for each new knowledge prototype:
\begin{equation}
\label{eq:kcm_att}
\begin{aligned}
\bm{k} = \delta(g_c(\widetilde{\bm{z}}^{t-1})),
\end{aligned}
\end{equation} where $g_c(\cdot)$ denotes a three-layer fully connected network, $\delta(\cdot)$ denotes the softmax function, and $(\widetilde{\,\cdot\,})$ denotes the l2-normalized function. Then, we combine the captured new knowledge with a weighted summation of the prototypes:
\begin{equation}
\label{eq:kcm_sum}
\begin{aligned}
\bm{z}^c = \sum_{p=1}^{P} \bm{k}^o_i \cdot \bm{v}_p.
\end{aligned}
\end{equation}
\par Sequentially, a parallel Forward Mapping Module (FMM) is designed to directly map the source feature to the target feature, thereby preserving the old knowledge from source domains. The designed mapping network consists of sequential fully connected layers, Batch Normalization~\cite{ioffe2015batch} layers, and PReLU~\cite{he2015delving} layers. The first fully connected layer reduces the channel number to $C_0$, while the last layer recovers it back to the original number, which keeps the representation capability while saving redundant operations:
\begin{equation}
\label{eq:fmm}
\begin{aligned}
\bm{z}^m = g_s(\widetilde{\bm{z}}^{t-1}).
\end{aligned}
\end{equation}
\par Finally, the BiCT network takes the input $\widetilde{\bm{z}}^{t-1}$ to generate a factor $a^m$ to adaptively balance the above two features ($\bm{z}^c$ and $\bm{z}^m$) obtained by the KCM module and FMM module. Meanwhile, the original features are added by the residual skip-connection to alleviate the gradient vanishing issue~\cite{wang2020unified}. The transferred process $\bm{\widehat{z}}^{t-1}$ of BiCT network can be formulated as follows:
\begin{equation}
\label{eq:cct}
\begin{aligned}
\bm{\widehat{z}}^{t-1} = (1-a^m)\cdot\bm{z}^c+a^m\cdot\bm{z}^m+\widetilde{\bm{z}}^{t-1}.
\end{aligned}
\end{equation}
\par Following~\cite{meng2021learning}, we leverage 4 cascaded BiCT networks to build up a strong transfer network, which sequentially transfers features from previous stages to the current stage. We transfer the source feature $\bm{z}^{t-1}$ and the target feature $\bm{z}^{t}$ to each other at the same time, and exploit the transferred source features $\bm{\widehat{z}}^{t-1}$ and target features $\bm{\widehat{z}}^{t}$ for subsequent continuous compatible representation learning.

\par Intuitively, the mapped old features $\bm{z}^m$ project the discrimination of old knowledge into the new feature space, while $\bm{z}^c$ conversely expresses the discriminative knowledge specific to the new data through independently learned new knowledge prototypes. Sequentially, $a^m$ can quantitatively control the balance between old and new knowledge. Therefore, to balance the old and new knowledge after incorporating new identity under different domain discrepancies, we designed a learnable $a^m$ that adaptively adjusts to achieve a more robust balance, which is evaluated in the subsequent experimental section.

\subsection{Bidirectional Compatible Distillation}

\par To make the transferred features compatible with the target features, we design a Bidirectional Compatible Distillation (BiCD) module to maintain the old and the new knowledge with each other by aligning the relationships of different people before and after feature transfer in both feature spaces. Specifically, given a mini-batch of training samples $\{x_{i}^{t},y_{i}^{t}\}^{B}_{i=1}$, where $B$ is the batch size, and a new feature $\bm{z}_i^t$ calculated by the new model $\phi(\cdot)^t$. We start by directly aligning the above features with l2-loss:
\begin{equation}
\label{eq:l_ca}
\begin{aligned}
\mathcal{L}_{bca} = -\displaystyle{\frac{1}{B}} \sum_{i=1}^{B}(\widetilde{\bm{z}}_i^t-\widetilde{\widehat{\bm{z}}}^{t-1}_i)^2.
\end{aligned}
\end{equation}
\par Although the above alignment loss promotes the compatibility of transferred features in the target feature space, it ignores the relationship knowledge within the source domain, thus limiting its ReID performance. Therefore, we propose a relationship distillation loss to overcome the above issue.

\par Firstly, we construct an affinity matrix $M^{t-1} \in \mathbb{R}^{B\times B}$ for stage $(t-1)^{th}$, which is calculated by the similarity between each pair of two features in $\{\bm{z}^{t-1}_i\}^B_{i=1}$ to represent the relationship between the corresponding images in the source feature space:
\begin{equation}
\label{eq:bcd_m}
\begin{aligned}
M^{t-1}_{i,j} = \displaystyle{\frac{e^{\langle\bm{z}^{t-1}_i,\bm{z}^{t-1}_j \rangle}}{\sum_{k=1}^{B}e^{\langle\bm{z}^{t-1}_i, \bm{z}^{t-1}_k\rangle}}}, (i, j\in [1,B]),
\end{aligned}
\end{equation} where $\langle\cdot,\cdot\rangle$ denotes the cosine similarity. Similarly, a transferred similarity matrix $M^{t}$ is also constructed to represent the relationship between the transferred features $\{\bm{z}^t_i\}^B_{i=1}$. Apparently, the alignment in~\ref{eq:l_ca} will promote higher similarity between samples with the same identity in $M^{t}$, which will conflict with the corresponding similarity in $M^{t-1}$. Therefore, we remove the similarities between samples belonging to the same identity to avoid the above conflicts for relationship distillation, where the affinity of different identities can be normalized as follows:
\begin{equation}
\label{eq:bcd_m0}
\begin{aligned}
\mathcal{\widehat{M}}^{x}_{i,j} = 
\begin{cases}
\frac{M^{x}_{i,j}}{\sum_{k\neq i}M^{x}_{i,k}}, ID(i)\neq ID(j)\\
0, \quad\,\,\,\, ID(i)=ID(j),
\end{cases}
\end{aligned}
\end{equation} where $ID(\cdot)$ denotes the identity of the given sample, and $x\in\{t-1,t\}$. Next, a widely-used knowledge distillation loss $\mathcal{L}_{cr}$ based on Kullback-Leibler (KL) divergence is imposed to distil the relationship between source features as follows:
\begin{equation}
\label{eq:bcd_kl}
\begin{aligned}
\mathcal{L}_{bcr} = -\displaystyle{\frac{1}{B}} \sum_{i=1}^{B} \widehat{M_{i}}^{t-1} log(\displaystyle{\frac{\widehat{M_{i}}^{t-1}}{\widehat{M_{i}}^{t}}}).
\end{aligned}
\end{equation}
\par In all, the loss function for the BiCD module can be calculated as follows:
\begin{equation}
\label{eq:bad_l_bcd}
\begin{aligned}
\mathcal{L}_{bcd} = \mu_1\mathcal{L}_{bca}+\mu_2\mathcal{L}_{bcr},
\end{aligned}
\end{equation} where $\mu_1$ and $\mu_2$ denote two hyper-parameters to balance our proposed bidirectional compatible distillation.

\subsection{Bidirectional Anti-forgetting Distillation}

\par Although the proposed BiCD module achieves compatibility by selectively distilling the relationship, it ignores the accumulated forgetting of old knowledge during the continual compatible transfer, thereby reducing the inter-class discrimination of old data. Additionally, the new discriminative knowledge suffers from the compatible distillation loss during the learning process of the BiCT network, limiting its further improvement over the old discriminative knowledge. Therefore, we propose a Bidirectional Anti-forgetting Distillation (BiAD) module to preserve the discriminative information within the old and new features in the updated feature space.

\par To this end, we employ the classifier head $\psi^{t-1}(\cdot)$ at stage $(t-1)^{th}$ to extract and distil the discriminative information. Therefore, we extract source identity logits $\bm{q}^{t-1}$ of the old feature $\bm{z}^{t-1}$ as the theoretical distribution of the discriminative information:
\begin{equation}
\label{eq:bad_old}
\begin{aligned}
\bm{q}^{t-1} = \delta(\psi^{t-1}(\bm{z}^{t-1})).
\end{aligned}
\end{equation}
\par Considering that the transferred feature $\bm{\widehat{z}}^{t-1}$ and $\bm{z}^{t-1}$ come from different domains, it is challenging to extract consistent discriminative information directly using $\psi^{t-1}(\cdot)$. To tackle the above problem, we impose the target distribution from old features to the transferred distribution to highlight the old knowledge within the transferred distribution, which can be formalized as follows:
\begin{equation}
\label{eq:bad_new}
\begin{aligned}
\bm{\widehat{q}}^{t-1} = \delta(\psi^{t-1}(\widetilde{\widehat{\bm{z}}}^{t-1}\cdot\bm{\sigma}^{t-1}+\bm{\mu}^{t-1})),
\end{aligned}
\end{equation} where $\bm{\sigma}^{t-1}$ and $\bm{\mu}^{t-1}$ denote the mean and variance calculated by features of the source domain. Sequentially, we design an anti-forgetting loss $\mathcal{L}_{ad}$ to preserve the old discriminative knowledge, which can be calculated as follows:
\begin{equation}
\label{eq:bad_l_ad}
\begin{aligned}
\mathcal{L}_{bad} = -\displaystyle{\frac{1}{B}} \sum_{i=1}^{B}\bm{q}_i^{t-1}log(\frac{\bm{q}_i^{t-1}}{\bm{\widehat{q}}_i^{t-1}}).
\end{aligned}
\end{equation}
\par Despite achieving the anti-forgetting, the distillation in \ref{eq:bad_l_ad} will also limit the discrimination of the transferred feature in the new feature space. Therefore, we introduce a discriminative consistency-based loss $\mathcal{L}_{dc}$ to balance the discrimination for both transferred and new features:
\begin{equation}
\label{eq:bad_l_dc}
\begin{aligned}
\mathcal{L}_{bdc} = -\displaystyle{\frac{1}{B}} \sum_{i=1}^{B} (1-\cos\langle\frac{\widetilde{\widehat{\bm{z}}}^{t-1}_i-\widetilde{\bm{z}_i}^{t-1}}{||\widetilde{\widehat{\bm{z}}}^{t-1}_i-\widetilde{\bm{z}_i}^{t-1}||_2},\frac{\widetilde{\bm{z}_i}^t-\widetilde{\bm{z}_i}^{t-1}}{||\widetilde{\bm{z}_i}^t-\widetilde{\bm{z}_i}^{t-1}||_2}\rangle).
\end{aligned}
\end{equation}
\par Therefore, the loss function for the BiAD module can be calculated as follows:
\begin{equation}
\label{eq:bad_l_bad}
\begin{aligned}
\mathcal{L}_{bad} = \mu_3\mathcal{L}_{bad}+\mu_4\mathcal{L}_{bdc},
\end{aligned}
\end{equation} where $\mu_3$ and $\mu_4$ denote two hyper-parameters when training the model.

\par \textbf{Discussion.} Inspired by~\cite{li2017learning}, our BiAD module employs the output of old tasks to resist identity forgetting. Specifically, let the old classifier $\psi^{t-1}(\cdot)$ be the old task, and let the bidirectional compatible transfer network be the new task. As the new task is learned, $\psi^{t-1}(\cdot)$ cannot retain strong old identity information as it did when learning the old task, leading to an increase in entropy and catastrophic forgetting of old knowledge. However, in the BiAD module, Eq.~\ref{eq:bad_l_ad} approximates the probability distribution of the old model's output through relative entropy, thus avoiding the entropy increase caused by the transfer of knowledge space in the old model and suppressing catastrophic forgetting of old identity information.

\subsection{Dynamic Feature Fusion}

\par Although the BiCT network achieves the compatible update from old gallery features to the new ones through bidirectional distillation, the diversity differences between different data domains prevent the consistent network from transferring old knowledge to new knowledge due to the imbalanced domain-specific knowledge. Specifically, when the discrepancy between two adjacent domains is relatively large, the BiCT network tends to sacrifice old knowledge in an effort to make the old features compatible with the new ones, where directly transferring the old features may lead to the catastrophic forgetting of the old knowledge. In contrast, when the difference is relatively small, the BiCT network benefits from retaining more old knowledge while limiting the learning of new knowledge due to the smaller compatibility difficulty. Therefore, it is necessary to adaptively update old features according to different domain discrepancies. Instead of fusing the model parameters, we propose a dynamic feature fusion module, which can dynamically weight the updated and pre-updated features according to the variety between different data domains, promoting the balance between knowledge capturing and domain compatibility.

\par Inspired by~\cite{xu2024lstkc}, we first obtain the difference $\epsilon_t$ between the predictions of the new and old models at the $t^{th}$ stage to access the degree of change in knowledge:
\begin{equation}
\label{eq:dff_diff}
\begin{aligned}
\epsilon^{t} = \frac{1}{N^t}\sum^{N^t}_{i=1}\sum^{N^t}_{j=1}|M^{t}_{i,j}-M^{t-1}_{i,j}|.
\end{aligned}
\end{equation}

\par Intuitively, $M_{t}$ measures the similarity between each image $x^t_i$ and all other images at the $t^{th}$ stage. Therefore, the average of all inter-stage differences can reflect the overall knowledge difference between both new and old domains. Then, we can obtain the updated gallery features $\mathcal{F}^{t}$ at the $t^{th}$ stage by merging the features extracted at the current stage and the dynamically transferred historical gallery features:
\begin{equation}
\label{eq:dff_gallery}
\begin{aligned}
\mathcal{F}^{t} &= \{\widehat{\phi}^t(\mathcal{G}^t),\epsilon^{t}\cdot\mathcal{F}^{t-1}+(1-\epsilon^{t})\cdot\theta^t(\mathcal{F}^{t-1})\},\\
\end{aligned}
\end{equation}
\begin{equation}
\label{eq:dff_ema}
\begin{aligned}
\widehat{\phi}^t&=\epsilon^{t}\cdot\phi^{t-1}+(1-\epsilon^{t})\phi^t,\\
\end{aligned}
\end{equation} where $\theta^t(\cdot)$ represent the trained BiCT network and ReID model at the $t^{th}$ stage. 

\par In this case, both the new features extracted by the new model and the updated old features can be compatible with each other in an adaptive manner by adjusting to the changes according to different domain gaps. Sequentially, the fused model $\widehat{\phi}^t$ will be served as $\phi^t$ when training at the $t+1^{th}$ stage.

\par \textbf{Discussion.} Our DDF module preserves old knowledge by continuously combining features from different temporal stages. As discussed in~\cite{xu2024lstkc}, larger distribution shifts typically lead to more severe parameter-level deviations as the model needs to adapt to new data distributions continuously. In particular, in the RFL-ReID task, although $\epsilon^{t}$ achieves anti-forgetting at the parameter level by fusing old and new models in Eq.~\ref{eq:dff_ema}, distribution shift still remains between query features extracted by the fused model and the transferred gallery features aligned with the new model after the updating. This is because the former is weighted by the old data distribution represented by the old model, while the latter only considers the distribution of new domain features. Therefore, we introduce this idea into the DDF module and suppress the potential distribution shift between features learned at different time periods by symmetrically fusing both old and new models, as well as the features before and after the updating at each stage.

\subsection{Objective Function}

\par In all, the objective function of our Bi-C\textsuperscript{2}R is formulated as follows:
\begin{equation}
\label{eq:bad_l_all}
\begin{aligned}
\mathcal{L} = \mu_1\mathcal{L}_{bca}+\mu_2\mathcal{L}_{bcr}+\mu_3\mathcal{L}_{bad}+\mu_4\mathcal{L}_{bdc}.
\end{aligned}
\end{equation}

\section{Theoretical Analysis and Justifications}

\emph{1) What is the influence of the Dynamic Feature Fusion module on knowledge anti-forgetting in lifelong re-identification?}

\par Taking the update of the historical gallery features from the $1^{st}$ stage to the $t^{th}$ stage ($t>1$) as an example, the continuously updated gallery feature $\mathcal{F}^t$ can be formalized as $\mathcal{F}^t=\theta^t(...\theta^{2}(\phi^1(\mathcal{G}^1))$. Let $\mathbb{E}^1_b$ and $\mathbb{E}^t_c$ be the theoretical error of the ReID model $\phi^1(\cdot)$ and the BiCT network $\theta^t(\cdot)$ at the $t^{th}$ stage, respectively. Accordingly, the theoretical error $\mathbb{E}_F^t$ without using the Dynamic Feature Fusion module can be formalized as follows:
\begin{equation}
\label{eq:dff_ed}
\begin{aligned}
\mathbb{E}^t_F=\mathbb{E}^t_c...\mathbb{E}^2_c\cdot\mathbb{E}^1_b,
\end{aligned}
\end{equation} where $\mathbb{E}^i_c\geq1$ due to the inevitable forgetting of old knowledge with error amplification of the old model in the lifelong learning process. Similarly, we can get the error $\mathbb{E}_D$ after using our proposed dynamic feature fusion module:
\begin{equation}
\label{eq:dff_ef}
\begin{aligned}
\mathbb{E}^t_D=\epsilon^{t}...(\epsilon^{2}\mathbb{E}^1_b+(1-\epsilon^{2})\mathbb{E}^2_c\mathbb{E}^1_b)+...+(1-\epsilon^{t})\mathbb{E}^t_c...\mathbb{E}^2_c\mathbb{E}^1_b).
\end{aligned}
\end{equation}
\par Therefore, by subtracting the above two errors, we can get the following result:
\begin{equation}
\label{eq:dff_ef_df}
\begin{aligned}
\mathbb{E}^t_F-\mathbb{E}^t_D=\sum_{i=2}^{t}\epsilon^{t}...\epsilon^{i}\cdot\mathbb{E}^{i-1}_c...\mathbb{E}^{1}_b\cdot(\mathbb{E}^{i}_c-1)\geq0.
\end{aligned}
\end{equation}
\par It is obvious that the theoretical error of our method $\mathbb{E}_F$ is smaller than the direct transfer strategy $\mathbb{E}_D$ when $\mathbb{E}^i_c\geq1$. Therefore, the final transferred feature is a weighted fusion of all previous features that pursue long-term knowledge accumulation.

\emph{2) Why $\epsilon^{t}$ is employed as the fusion coefficient instead of others?}

\par Taking our baseline framework~\cite{xu2024lstkc} as the example, when learning at stage $t^{th}$, the parameter fusion weight $\epsilon^{t}$ is calculated by Eq.~\ref{eq:dff_diff}, while the fused model $\phi^t_q$ is served as the feature extraction model for the query image $\mathcal{Q}^t$ at the $t^{th}$ stage, which can be calculated as follows:
\begin{equation}
\label{eq:dff_param_fuse}
\begin{aligned}
\phi^t_q(\mathcal{Q}^t)=(\epsilon^{t}\cdot\phi^{t-1}+(1-\epsilon^{t})\cdot\phi^{t})(\mathcal{Q}^t),
\end{aligned}
\end{equation}
\par Meanwhile, the feature extraction and updating process of gallery images at the $t^{th}$ stage can be obtained by combining the old gallery feature $\mathcal{F}_{t-1}$:
\begin{equation}
\label{eq:dff_gallery_ext}
\begin{aligned}
\mathcal{F}^t=\left\{
\begin{array}{ll}
\epsilon^{t}\cdot\mathcal{F}^{t-1}+(1-\epsilon^{t})\cdot\theta^t(\mathcal{F}^{t-1})\\
(\epsilon^{t}\cdot\phi^{t-1}+(1-\epsilon^{t})\cdot\phi^{t})(\mathcal{G}^{t})
\end{array}
\right.
\end{aligned}
\end{equation}
\par Therefore, to achieve the compatibility between the query features $\phi_q^t(x)$ obtained by the fused latest inference model and the gallery features $\mathcal{F}^t$ extracted and updated by the old model, we use the same feature and model fusion weights $\epsilon^{t}$ that assures a more stable long-term re-indexing knowledge accumulation.

Specifically, given a feature updating network $\theta^t(\mathcal{F}^{t-1})$ after the $t^{th}$ stage training phase, which is employed to approximate $\phi^{t}(x^{t-1})$, let $\alpha_1$ and $\alpha_2$ be two fusion weights in Eq.~\ref{eq:dff_param_fuse} and feature fusion in Eq.~\ref{eq:dff_gallery_ext}, respectively:
\begin{equation}
\label{eq:dff_param_fuse_a1}
\begin{aligned}
\phi^t_q(X)=(\alpha_1\cdot\phi^{t-1}+(1-\alpha_1)\cdot\phi^{t})(X),
\end{aligned}
\end{equation}
\begin{equation}
\label{eq:dff_gallery_ext_a2}
\begin{aligned}
\mathcal{F}^t=\alpha_2\cdot\mathcal{F}^{t-1}+(1-\alpha_2)\cdot\theta^t(\mathcal{F}^{t-1}).
\end{aligned}
\end{equation}
\par Therefore, the difference between the query and the gallery features can be described as:
\begin{equation}
\label{eq:dff_cha}
\begin{aligned}
\mathbb{E}^t_{a_1,a_2}\sim
|(\alpha_1-\alpha_2)\cdot\mathcal{F}^{t-1}+(\alpha_2-\alpha_1)\cdot\mathcal{F}^{t}|+C_{t},
\end{aligned}
\end{equation} where $C_{t}$ denote the constant discrepancy derived from network fusing. It can be seen that the optimal accumulated difference $\mathbb{E}^t_{a_1,a_2}$ is achieved when $\alpha_1=\alpha_2=\epsilon^t$. Therefore, inspired by~\cite{xu2024lstkc}, the adopted fusion weight $\epsilon^t$ in Eq.~\ref{eq:dff_cha} achieves compatible long-term knowledge accumulation by adaptively balancing the differences between the old and new domains.

\emph{3) What is the influence of bidirectional compatible transfer over forward transfer?}

Different from the existing L-ReID methods, the optimization objective of our Bi-C\textsuperscript{2}R can be described as the distance $F^t$ between samples at the $t^{th}$ stage:
\begin{equation}
\label{eq:bct_obj}
\begin{aligned}
F^t=&F_{(q^t,p^t)\in \mathcal{D}^t}\sim(\phi^t(q^t),\phi^t(p^t))\\&+F_{(q^s,p^s)\in \mathcal{D}^s}\sim(\phi^t(q^s),\mathcal{T}^{s\rightarrow t}\circ\phi^s(p^s)),
\end{aligned}
\end{equation} where the first expression denotes the learning objective of the new model on new data, which is optimized within the current stage $t$, and the second expression denotes the learning objective between the updated gallery features and the current query features, where its upper boundary $\varepsilon^t$ can be further derived from the triangular inequality:
\begin{equation}
\label{eq:bct_obj_tri}
\begin{aligned}
\varepsilon^t\leq&F_{(q^s,p^s)\in \mathcal{D}^s}\sim(\phi^t(q^s),\phi^t(p^s))\\&+F_{(q^s,p^s)\in \mathcal{D}^s}\sim(\phi^t(p^s),\mathcal{T}^{s\rightarrow t}\circ\phi^s(p^s)).
\end{aligned}
\end{equation}

Obviously, the expression $F_{(q^s,p^s)\in \mathcal{D}^s}\sim(\phi^t(q^s),\phi^t(p^s))$ requires that the old data in stage $s$ can maintain its discriminability in the new feature space (stage $t$).

However, due to the unavailability of old data, the discriminative information of the new data extracted by the old model in the forward transfer (as in Eq.~\ref{eq:bad_old}) suffers from the discrepancy between the old and new domain, resulting in insufficient distillation of discriminative information. In contrast, our bidirectional compatible transfer method exploits the new discriminative information extracted by the new model to reconstruct and align the old discriminative information in the old feature space (as in Eq.~\ref{eq:bad_l_dc} and Eq.~\ref{eq:bad_l_ad}), thereby mitigating the domain discrepancy and improving the discriminability of the transferred features.

In addition, the expression $F_{(q^s,p^s)\in \mathcal{D}^s}\sim(\phi^t(p^s),\mathcal{T}^{s\rightarrow t}\circ\phi^s(p^s))$ requires that the transferred features are aligned with the new features for the same instance in the old domain. According to~\cite{xu2024distribution}, the discrepancy between the old and new features mainly derived from the domain gap, regarding diverse domain styles, \emph{e.g.,} hue and saturation. It can be seen that forward transfer compromises the old data by employing new data to train the feature transfer network, where the transferred feature ignores the discriminativeness of the old knowledge. In contrast, the bidirectional transfer method preserves the mutual consistency of the mapping between the new and old domains by further introducing the backward transfer process from new features to old features. It not only improves the consistency of the instance distribution between the new and old feature spaces  (Eq.~\ref{eq:bcd_kl}), but also facilitates the compatibility between the new discriminative knowledge and the old discriminative knowledge (Eq.~\ref{eq:l_ca}), thereby further enhancing the consistent compatibility of the transferred features.

In summary, our bidirectional compatible transfer possesses a smaller theoretical risk than the forward transfer.

\begin{table*}[htbp]
\caption{Performance on training Order-1: Market1501$\rightarrow$CUHK-SYSU$\rightarrow$DukeMTMC$\rightarrow$MSMT17$\rightarrow$CUHK03.}
\renewcommand{\arraystretch}{0.92}
\normalsize
  \centering
  \setlength{\tabcolsep}{1mm}{
    \begin{tabular}{c|l|c|cc|cc|cc|cc|cc|cc}
    \hline
    \multirow{2}[1]{*}{Task}&\multicolumn{1}{c|}{\multirow{2}[1]{*}{Method}}&\multirow{2}[1]{*}{Venue}
   & \multicolumn{2}{c|}{Market1501} & \multicolumn{2}{c|}{CUHK-SYSU} & \multicolumn{2}{c|}{DukeMTMC} & \multicolumn{2}{c|}{MSMT17} & \multicolumn{2}{c}{CUHK03} & \multicolumn{2}{|c}{\textbf{Average}} \\
    \cline{4-15}          & & & mAP   & R1   & mAP   & R1   & mAP   & R1   & mAP   & R1   & mAP   & R1   & mAP   & R1 \\
    \hline
    \multicolumn{2}{c|}{JointTrain}&-& 68.1 &85.2 &81.4 &83.8 &60.4 &75.7 &24.6 &48.9 &42.7 &43.6 &55.4 &67.5\\
    \hline
    \multirow{10}[2]{*}{\rotatebox[origin=c]{90}{L-ReID}}
    &SPD~\cite{tung2019similarity}   &\emph{\color{gray}{ICCV 2019}}&35.6 &61.2 &61.7 &64.0 &27.5 &47.1 &5.2 &15.5 &42.2 &44.3 &34.4 &46.4 \\
    &LwF~\cite{li2017learning}       &\emph{\color{gray}{T-PAMI 2017}}&56.3 &77.1 &72.9 &75.1 &29.6 &46.5 &6.0 &16.6 &36.1 &37.5 &40.2 &50.6 \\
    &CRL~\cite{zhao2021continual}    &\emph{\color{gray}{WACV 2021}}&58.0 &78.2 &72.5 &75.1 &28.3 &45.2 &6.0 &15.8 &37.4 &39.8 &40.5 &50.8 \\
    &AKA~\cite{pu2021lifelong}      &\emph{\color{gray}{CVPR 2021}}& 58.1 &77.4 &72.5 &74.8 &28.7 &45.2 &6.1 &16.2 &38.7 &40.4 &40.8 &50.8 \\
    &MEGE~\cite{pu2023memorizing}    &\emph{\color{gray}{T-PAMI 2023}}& 39.0    & 61.6  & 73.3  & 76.6  & 16.9  & 30.3  & 4.6   & 13.4  & 36.4  & 37.1  & 34.0    & 43.8  \\
    &PatchKD~\cite{sun2022patch}     &\emph{\color{gray}{ACM MM 2022}}& 68.5 & 85.7 & 75.6 & 78.6 & 33.8 & 50.4 &6.5 & 17.0 &34.1 &36.8 & 43.7 & 53.7 \\
    &LSTKC~\cite{xu2024lstkc}   &\emph{\color{gray}{AAAI 2024}} &54.7 &76.0 &81.1 &83.4 &49.4 &66.2 &20.0 &43.2 &44.7 &46.5 &50.0 &63.1\\
    &DKP~\cite{xu2024distribution} &\emph{\color{gray}{CVPR 2024}} &60.3 &80.6 &83.6 &85.4 &51.6 &68.4 &19.7 &41.8 &43.6 &44.2 &\textcolor{red}{\textbf{51.8}} &\textcolor{blue}{\textbf{64.1}}\\
    \cline{2-15}
    \rowcolor{gray!15}\cellcolor{gray!0} & C\textsuperscript{2}R\textsuperscript{*}~\cite{cui2024learning} &\emph{\color{gray}{CVPR 2024}} &57.5 	&77.7 	&82.9 	&84.1 	&50.9 	&67.4 	&20.2 	&44.0 	&42.9 	&43.7 	&50.9 	&63.4\\
    \rowcolor{gray!15}\cellcolor{gray!0}& Bi-C\textsuperscript{2}R & This Paper &56.4	&77.6	&82.8	&84.8	&52.5	&69.6	&21.6	&46.3	&43.6	&44.5	&\textcolor{blue}{\textbf{51.4}} & \textcolor{red}{\textbf{64.6}}\\
    \hline
    \multirow{8}[2]{*}{\rotatebox[origin=c]{90}{\ \ RFL-ReID}}
    &LwF~\cite{li2017learning}       &\emph{\color{gray}{T-PAMI 2017}} &39.1 &58.0 &40.0 &40.7 &7.8 &15.3 &2.6 &7.1 &23.3 &23.9 &22.6 &29.0\\
    &AKA~\cite{pu2021lifelong}      &\emph{\color{gray}{CVPR 2021}} &36.1 &52.2 &38.6 &37.6 &7.6 &13.8 &3.1 &8.3 &26.5 &26.5 &22.4 &27.7\\
    &CVS~\cite{wan2022continual}     &\emph{\color{gray}{CVPR 2022}} &38.8 &55.6 &49.0 &49.7 &19.3 &30.0 &4.6 &11.5 &24.7 &24.7 &27.3 &34.3 \\
    &PatchKD~\cite{sun2022patch}     &\emph{\color{gray}{ACM MM 2022}}& 61.4 & 78.4 & 57.8 & 59.0 & 20.8 & 34.4 & 5.1 & 12.8 & 36.0 & 37.6 & 36.2 & 44.4 \\
    &LSTKC~\cite{xu2024lstkc} &\emph{\color{gray}{AAAI 2024}}&49.8 	&66.0 	&73.4 	&74.9 	&47.2 	&64.0 	&21.2 	&43.5 	&43.9 	&44.8 	&\textcolor{blue}{\textbf{47.1}} 	&\textcolor{blue}{\textbf{58.6}} \\
    &DKP~\cite{xu2024distribution} &\emph{\color{gray}{CVPR 2024}} &35.8	&52.5	&50.0	&49.6	&38.1	&55.9	&17.7	&37.4	&42.9	&44.0	&36.9	&47.9\\
    \cline{2-15}
    \rowcolor{gray!15}\cellcolor{gray!0} & C\textsuperscript{2}R\textsuperscript{*}~\cite{cui2024learning} &\emph{\color{gray}{CVPR 2024}} &54.4 	&70.5 	&75.0 	&76.8 	&46.2 	&64.0 	&19.6 	&42.1 	&42.9 	&43.7  &47.6 	&59.4 \\
    \rowcolor{gray!15}\cellcolor{gray!0}& Bi-C\textsuperscript{2}R & This Paper &57.1	&71.2	&75.1	&76.4	&51.3	&67.8	&23.3	&47.4	&43.6	&44.5	&\textcolor{red}{\textbf{50.1}}	&\textcolor{red}{\textbf{61.5}}\\
    
    \hline
    \end{tabular}
    }
  \label{tab: setting1}
\end{table*}

\begin{table*}[htbp]

\caption{Performance on training Order-2: DukeMTMC$\rightarrow$MSMT17$\rightarrow$Market1501$\rightarrow$CUHK-SYSU$\rightarrow$CUHK03.}

\renewcommand{\arraystretch}{0.92}
\normalsize
  \centering
  \setlength{\tabcolsep}{1mm}{
    \begin{tabular}{c|l|c|cc|cc|cc|cc|cc|cc}
    \hline
    \multirow{2}[1]{*}{Task}&\multicolumn{1}{c|}{\multirow{2}[1]{*}{Method}}&\multirow{2}[1]{*}{Venue}
   & \multicolumn{2}{c|}{DukeMTMC} & \multicolumn{2}{c|}{MSMT17} & \multicolumn{2}{c|}{Market1501} & \multicolumn{2}{c|}{CUHK-SYSU} & \multicolumn{2}{c}{CUHK03} & \multicolumn{2}{|c}{\textbf{Average}} \\
    \cline{4-15}          & & & mAP   & R1   & mAP   & R1   & mAP   & R1   & mAP   & R1   & mAP   & R1   & mAP   & R1 \\
    \hline
    \multicolumn{2}{c|}{JointTrain}&-&60.4 &75.7 &24.6 &48.9 &68.1 &85.2 &81.4 &83.8 &42.7 &43.6 &55.4 &67.5\\
    \hline
    \multirow{10}[2]{*}{\rotatebox[origin=c]{90}{L-ReID}}
    &SPD~\cite{tung2019similarity}   &\emph{\color{gray}{ICCV 2019}} &28.5 &48.5 &3.7 &11.5 &32.3 &57.4 &62.1 &65.0 &43.0 &45.2 &33.9 &45.5 \\
    &LwF~\cite{li2017learning}       &\emph{\color{gray}{T-PAMI 2017}}&42.7 &61.7 &5.1 &14.3 &34.4 &58.6 &69.9 &73.0 &34.1 &34.1 &37.2 &48.4 \\
    &CRL~\cite{zhao2021continual}    &\emph{\color{gray}{WACV 2021}}&43.5 &63.1 &4.8 &13.7 &35.0 &59.8 &70.0 &72.8 &34.5 &36.8 &37.6 &49.2 \\
    &AKA~\cite{pu2021lifelong}      &\emph{\color{gray}{CVPR 2021}}&42.2 &60.1 &5.4 &15.1 &37.2 &59.8 &71.2 &73.9 &36.9 &37.9 &38.6 &49.4 \\
    &MEGE~\cite{pu2023memorizing}    &\emph{\color{gray}{T-PAMI 2023}}&21.6 &35.5 &3.0 &9.3 &25.0 &49.8 &69.9 &73.1 &34.7 &35.1 &30.8 &40.6 \\
    &PatchKD~\cite{sun2022patch}     &\emph{\color{gray}{ACM MM 2022}}&58.3 &74.1 &6.4 &17.4 &43.2 &67.4 &74.5 &76.9 &33.7 &34.8 &43.2 &54.1\\
    &LSTKC~\cite{xu2024lstkc}   &\emph{\color{gray}{AAAI 2024}} &49.9 &67.6 &14.6 &34.0 &55.1 &76.7 &82.3 &83.8 &46.3 &48.1 &49.6 &62.1\\
    &DKP~\cite{xu2024distribution} &\emph{\color{gray}{CVPR 2024}} &53.4 &70.5 &14.5 &33.3 &60.6 &81.0 &83.0 &84.9 &45.0 &46.1 &\textcolor{blue}{\textbf{51.3}} &\textcolor{blue}{\textbf{63.2}}\\
    \cline{2-15}
    \rowcolor{gray!15}\cellcolor{gray!0} & C\textsuperscript{2}R\textsuperscript{*}~\cite{cui2024learning} &\emph{\color{gray}{CVPR 2024}} & 54.7	& 71.7	& 17.8	& 39.4	& 56.2	& 77.9	& 84.1	& 86.1	& 40.9	& 41.6	& 50.7	& 63.4	\\
    \rowcolor{gray!15}\cellcolor{gray!0}& Bi-C\textsuperscript{2}R & This Paper&54.0	&70.4	&17.2	&38.4	&57.5	&79.5	&84.2	&85.8	&45.0	&47.9	&\textcolor{red}{\textbf{51.6}}	&\textcolor{red}{\textbf{64.4}}\\
    \hline
    \multirow{8}[2]{*}{\rotatebox[origin=c]{90}{\ \ RFL-ReID}}
    &LwF~\cite{li2017learning}       &\emph{\color{gray}{T-PAMI 2017}} &15.0 &22.9 &1.2 &3.2 &9.5 &19.4 &38.8 &37.5 &20.2 &19.6 &16.9 &20.5\\
    &AKA~\cite{pu2021lifelong}      &\emph{\color{gray}{CVPR 2021}} &11.1 &15.1 &1.3 &3.2 &13.4 &27.3 &35.9 &34.7 &25.2 &25.6 &17.4 &21.2\\
    &CVS~\cite{wan2022continual}     &\emph{\color{gray}{CVPR 2022}} &29.0 &41.9 &3.5 &9.4 &30.7 &49.6 &60.0 &61.2 &28.5 &29.9 &30.3 &38.4 \\
    &PatchKD~\cite{sun2022patch}     &\emph{\color{gray}{ACM MM 2022}}&46.5 &60.9 &4.0 &10.4 &31.1 &50.5 &63.0 &64.0 &35.8 &36.6 &36.1 &44.5 \\
    
    &LSTKC~\cite{xu2024lstkc} &\emph{\color{gray}{AAAI 2024}}&41.0	&56.1	&16.3	&35.4	&56.0	&76.4	&83.3	&84.7	&45.8	&46.8	&\textcolor{blue}{\textbf{48.5}}	&\textcolor{blue}{\textbf{59.9}}\\

    &DKP~\cite{xu2024distribution} &\emph{\color{gray}{CVPR 2024}}&34.4	&50.4	&10.9	&25.1	&51.9	&73.0	&80.9	&82.3	&43.8	&44.9	&44.4	&55.2\\

    \cline{2-15}
    \rowcolor{gray!15}\cellcolor{gray!0} & C\textsuperscript{2}R\textsuperscript{*}~\cite{cui2024learning} &\emph{\color{gray}{CVPR 2024}} & 51.1	& 66.5	& 17.7	& 37.6	& 53.7	& 75.7	& 82.9	& 84.2	& 40.9	& 41.6	& 49.3	& 61.1\\
    \rowcolor{gray!15}\cellcolor{gray!0}& Bi-C\textsuperscript{2}R & This Paper &51.6	&66.4	&18.7	&38.6	&56.7	&77.6	&83.9	&85.4	&45.0	&47.9	&\textcolor{red}{\textbf{51.2}}	&\textcolor{red}{\textbf{63.2}}\\
    \hline
    \end{tabular}
    }
  \label{tab: setting2}
\end{table*}

\emph{4) How does the proposed architecture balance new knowledge acquisition with old knowledge preservation?}

As discussed in~\cite{xu2024distribution}, the identity distribution can be modeled as two Gaussian distributions $p(X^t)\sim\mathcal{N}(\mathbb{\mu}^t,\mathbb{\sigma}^t)$, $p(X^{t-1})\sim\mathcal{N}(\mathbb{\mu}^{t-1},\mathbb{\sigma}^{t-1})$ for both the new data $X^t$ and old data $X^{t-1}$. Therefore, the discrepancy between the new and the old distribution can be described as follows:
\begin{equation}
\label{eq:trans}
\begin{aligned}
X^t=\frac{\mathbb{\sigma}^t}{\mathbb{\sigma}^{t-1}}X^{t-1}-\frac{\mathbb{\sigma}^t}{\mathbb{\sigma}^{t-1}}\mathbb{\mu}^{t-1}+\mathbb{\mu}^t.
\end{aligned}
\end{equation}
\par Based on the above preliminary, the process of acquiring new knowledge and preserving old knowledge can be correlated with our designed bidirectional transfer architecture. We begin with the transfer from old features to new features. Intuitively, the Forward Mapping Module (FFM) in our bidirectional compatible transfer network captures new knowledge by mapping the discriminative reliability $\mathbb{\sigma}^{t-1}$ of the old identity distribution to the new feature space, thus addressing the variance difference $\frac{\mathbb{\sigma}^t}{\mathbb{\sigma}^{t-1}}$ between old and new knowledge, employing stacked linear mapping layers, activation function, etc. Meanwhile, the Knowledge Capturing Module (KCM) compensates for the relative difference $-\frac{\mathbb{\sigma}^t}{\mathbb{\sigma}^{t-1}}\mathbb{\mu}^{t-1}+\mathbb{\mu}^t$ that are unrelated to the old data $X^{t-1}$ for high-dimensional distribution mean between new and old knowledge by weighting a set of knowledge prototypes $\bm{v}_p$. Similarly, the knowledge reconstruction from new features to old features can be achieved by the reverse process of Eq.~\ref{eq:trans}, thus ensuring the retention of old knowledge. By simultaneously balancing the bidirectional transfer and knowledge distillation between new and old knowledge, our architecture can achieve a better balance between both knowledge.

\begin{table*}[!t]
\renewcommand{\arraystretch}{1.05}
  \caption{Performance on training Order-8: Market$\rightarrow$LTCC$\rightarrow$PRCC$\rightarrow$MSMT17$\rightarrow$CUHK03.}
  \centering
  \small
  \setlength{\tabcolsep}{0.65mm}{
    \begin{tabular}{c|l|cc|cc|cc|cc|cc||cc|cc||cc}
    \hline
    & & \multicolumn{10}{c||}{Cloth-Consistent}& \multicolumn{4}{c||}{Cloth-Changing}& \multicolumn{2}{c}{\multirow{2}{*}{Average}}\\
    \cline{3-16}
    \multicolumn{1}{c|}{Task} & \multicolumn{1}{c|}{Method} & \multicolumn{2}{c|}{Market} & \multicolumn{2}{c|}{LTCC} & \multicolumn{2}{c|}{PRCC} & \multicolumn{2}{c|}{MSMT17} & \multicolumn{2}{c||}{CUHK03} & \multicolumn{2}{c|}{LTCC} & \multicolumn{2}{c||}{PRCC} & & \\
    \cline{3-18}
    & & mAP   & R1   & mAP   & R1   & mAP   & R1   & mAP   & R1   & mAP   & R1   & mAP   & R1   & mAP   & R1   & mAP   & R1\\
    \hline
    \multicolumn{2}{c|}{JointTrain}           & 64.1 	&82.5 	&42.6 	&62.1 	&94.6 	&98.7 	&18.4 	&40.8 	&44.4 	&46.4 	&10.1 	&23.0 	&32.7 	&33.8 	&43.8 	&55.3\\
    \hline
    \multirow{2}{*}{L-ReID}&LSTKC &39.9 	&63.4 	&39.6 	&65.4 	&95.9 	&98.9 	&11.5 	&29.2 	&48.1 	&50.1 	&8.3 	&19.4 	&24.0 	&22.9 	&38.2 	&49.9\\
    \cline{2-18}
    & \cellcolor{gray!15}Bi-C\textsuperscript{2}R & \cellcolor{gray!15}55.2 & \cellcolor{gray!15}76.2 & \cellcolor{gray!15}52.6 & \cellcolor{gray!15}73.6 & \cellcolor{gray!15}97.0 & \cellcolor{gray!15}98.4 & \cellcolor{gray!15}20.2 & \cellcolor{gray!15}44.3 & \cellcolor{gray!15}42.2 & \cellcolor{gray!15}43.4 & \cellcolor{gray!15}10.0 & \cellcolor{gray!15}22.2 & \cellcolor{gray!15}26.7 & \cellcolor{gray!15}23.8 & \cellcolor{gray!15}\textbf{43.1} & \cellcolor{gray!15}\textbf{54.6}\\
    \hline
    \multirow{2}{*}{RFL-ReID}&LSTKC &32.0 	&49.4 	&37.6 	&61.7 	&91.3 	&96.3 	&12.8 	&30.8 	&48.1 	&50.1 	&8.3 	&18.7 	&20.4 	&20.3 	&35.8 	&46.8\\
    \cline{2-18}
    & \cellcolor{gray!15}Bi-C\textsuperscript{2}R & \cellcolor{gray!15}56.0 & \cellcolor{gray!15}70.4 & \cellcolor{gray!15}52.8 & \cellcolor{gray!15}72.9 & \cellcolor{gray!15}93.6 & \cellcolor{gray!15}96.7 & \cellcolor{gray!15}21.6 & \cellcolor{gray!15}45.0 & \cellcolor{gray!15}42.2 & \cellcolor{gray!15}43.4 & \cellcolor{gray!15}10.0 & \cellcolor{gray!15}19.1 & \cellcolor{gray!15}23.4 & \cellcolor{gray!15}22.9 & \cellcolor{gray!15}\textbf{42.8} & \cellcolor{gray!15}\textbf{52.9}\\
    \hline
    \end{tabular}%
    }
    \vspace{-5pt}
  \label{tab: setting8}%
\end{table*}%

\begin{table*}[htbp]
\renewcommand{\arraystretch}{1.05}
  \caption{Performance on training Order-9: MSMT17$\rightarrow$PRCC$\rightarrow$Market$\rightarrow$CUHK03$\rightarrow$LTCC.}
  \centering
  \small
  \setlength{\tabcolsep}{0.65mm}{
    \begin{tabular}{c|l|cc|cc|cc|cc|cc||cc|cc||cc}
    \hline
    & & \multicolumn{10}{c||}{Cloth-Consistent}& \multicolumn{4}{c||}{Cloth-Changing}& \multicolumn{2}{c}{\multirow{2}{*}{Average}}\\
    \cline{3-16}
    \multicolumn{1}{c|}{Task} & \multicolumn{1}{c|}{Method} & \multicolumn{2}{c|}{MSMT17} & \multicolumn{2}{c|}{PRCC} & \multicolumn{2}{c|}{Market} & \multicolumn{2}{c|}{CUHK03} & \multicolumn{2}{c||}{LTCC} & \multicolumn{2}{c|}{PRCC} & \multicolumn{2}{c||}{LTCC} & & \\
    \cline{3-18}
    & & mAP   & R1   & mAP   & R1   & mAP   & R1   & mAP   & R1   & mAP   & R1   & mAP   & R1   & mAP   & R1   & mAP   & R1\\
    \hline
    \multicolumn{2}{c|}{JointTrain}           &18.4 	&40.8 	&94.6 	&98.7 	&64.1 	&82.5 	&44.4 	&46.4 	&42.6 	&62.1 	&32.7 	&33.8 	&10.1 	&23.0 	&43.8 	&55.3\\
    \hline
    \multirow{2}{*}{L-ReID}&LSTKC &6.9	&20.4	&92.8	&97.0	&39.8	&64.3	&25.2	&26.0	&51.1	&65.5 	&27.3	&27.0	&10.9	&22.4 	&36.3 	&46.1\\
    \cline{2-18}
    & \cellcolor{gray!15}Bi-C\textsuperscript{2}R & \cellcolor{gray!15}17.4 & \cellcolor{gray!15}39.3 & \cellcolor{gray!15}96.3 & \cellcolor{gray!15}99.3 & \cellcolor{gray!15}47.1 & \cellcolor{gray!15}70.9 & \cellcolor{gray!15}31.8 & \cellcolor{gray!15}32.3 & \cellcolor{gray!15}58.9 & \cellcolor{gray!15}76.5 & \cellcolor{gray!15}29.7 & \cellcolor{gray!15}27.9 & \cellcolor{gray!15}12.2 & \cellcolor{gray!15}23.5 & \cellcolor{gray!15}\textbf{41.9} & \cellcolor{gray!15}\textbf{52.8}\\
    \hline
    \multirow{2}{*}{RFL-ReID}&LSTKC &3.6	&10.2	&71.0	&86.4	&33.7	&55.8	&27.4	&27.7	&51.1	&65.5 	&15.0	&16.2	&10.9	&22.4 	&30.4 	&40.6\\
    \cline{2-18}
    & \cellcolor{gray!15}Bi-C\textsuperscript{2}R & \cellcolor{gray!15}16.6 & \cellcolor{gray!15}33.7 & \cellcolor{gray!15}76.6 & \cellcolor{gray!15}88.1 & \cellcolor{gray!15}43.9 & \cellcolor{gray!15}65.1 & \cellcolor{gray!15}34.5 & \cellcolor{gray!15}35.2 & \cellcolor{gray!15}58.9 & \cellcolor{gray!15}76.5 & \cellcolor{gray!15}18.4 & \cellcolor{gray!15}18.3 & \cellcolor{gray!15}12.2 & \cellcolor{gray!15}23.5 & \cellcolor{gray!15}\textbf{37.3} & \cellcolor{gray!15}\textbf{48.6}\\
    \hline
    \end{tabular}%
    }
    \vspace{-5pt}
  \label{tab: setting9}%
\end{table*}%

\section{Experiments}

\subsection{Datasets and Evaluation Metrics}

In this section, we introduce the benchmark datasets and evaluation metrics used to verify the effectiveness of our method.

\noindent\textbf{Datasets.} To verify the effectiveness of our proposed Bi-C\textsuperscript{2}R, seven person ReID datasets, \emph{i.e.,} Market1501~\cite{market1501}, CUHK-SYSU~\cite{cuhksysu}, DukeMTMC~\cite{dukemtmc}, MSMT17~\cite{msmt17v2}, CUHK03~\cite{cuhk03}, LTCC~\cite{qian2020long}, and PRCC~\cite{yang2019person} are employed to evaluate the performance under different scenarios. Specifically, two benchmarking orders~\cite{pu2021lifelong} are conducted to perform verify the robustness, \emph{i.e.,} Order-1: Market1501$\rightarrow$CUHK-SYSU$\rightarrow$DukeMTMC$\rightarrow$MSMT17$\rightarrow$ CUHK03, Order-2: DukeMTMC$\rightarrow$MSMT17$\rightarrow$Market1501$\rightarrow$ CUHK-SYSU$\rightarrow$CUHK03. Besides, another 5 different orders (\emph{i.e.,} Order-3: MSMT17$\rightarrow$CUHK-SYSU$\rightarrow$DukeMTMC $\rightarrow$Market1501$\rightarrow$CUHK03, Order-4: DukeMTMC$\rightarrow$Market1501$\rightarrow$CUHK03$\rightarrow$MSMT17$\rightarrow$CUHK-SYSU, Order-5: CUHK-SYSU$\rightarrow$DukeMTMC$\rightarrow$CUHK03 $\rightarrow$MSMT17$\rightarrow$Market1501, Order-6: CUHK03$\rightarrow$MSMT17$\rightarrow$ DukeMTMC$\rightarrow$Market1501$\rightarrow$CUHK-SYSU, and Order-7: CUHK03$\rightarrow$MSMT17$\rightarrow$DukeMTMC$\rightarrow$Market1501$\rightarrow$CUHK-SYSU) in~\cite{pu2023memorizing} are also introduced to verify the anti-forgetting and generalization ability under more variation. Moreover, two additional orders are employed to verify the generalizability under more practical scenarios, including Order-8: Market1501$\rightarrow$LTCC$\rightarrow$PRCC$\rightarrow$MSMT17$\rightarrow$CUHK03, Order-9: MSMT17$\rightarrow$PRCC$\rightarrow$Market1501$\rightarrow$CUHK03$\rightarrow$LTCC, where LTCC and PRCC contain more camera configurations, lighting conditions, and clothing variation in real-world deployment scenarios.

\noindent\textbf{Evaluation Metrics.} Two ReID metrics (mean Average Precision (mAP)~\cite{market1501} and R1 accuracy (R1)~\cite{moon2001computational}) are adopted to compare our method with other methods and evaluate the effectiveness of each component on each dataset. In addition, the Average Forgetting (AF)~\cite{chaudhry2018riemannian} performance of the above metrics is employed for the comparison of the anti-forgetting performance after sequential training on different orders, averaging the performance degradation on each dataset across the lifelong re-identification process.

\subsection{Implementation Details}
We use Pytorch~\cite{paszke2019pytorch} to implement our Bi-C\textsuperscript{2}R on a single NVIDIA A40 GPU. The pre-training ResNet-50 network~\cite{he2016deep} on ImageNet~\cite{russakovsky2015imagenet} dataset is employed as our backbone. We use LSTKC~\cite{xu2024lstkc} as the baseline of our method. Following LSTKC, we train our Bi-C\textsuperscript{2}R for 80 epochs on the first stage and 60 epochs for the others with SGD optimizer, where the learning rate is set to $8\times10^{-3}$ initially with 0.1 weight decay at the 30$^{th}$ epoch within each training stage. Each input images are resized to 256$\times$128 with random erasing, cropping, and horizontal flipping for data augmentation. The prototype length $P$ and the channel number $C_0$ are empirically set to 16 and 32. The batch size is set to 64 with 16 identities and 4 randomly sampled images for each identity. The hyper-parameters $\mu_1$, $\mu_2$, $\mu_3$, and $\mu_4$ are set to $100$, $1$, $7e^{-2}$ and $5^{e-4}$, respectively.

\subsection{Comparison with State-of-the-arts Methods}
In this section, we compared our proposed Bi-C\textsuperscript{2}R with several state-of-the-art LReID methods on both traditional L-ReID and our RFL-ReID tasks, including SPD~\cite{tung2019similarity}, LwF~\cite{li2017learning}, CRL~\cite{zhao2021continual}, AKA~\cite{pu2021lifelong}, MEGE~\cite{pu2023memorizing}, PatchKD~\cite{sun2022patch}, LSTKC~\cite{xu2024lstkc}, and DKP~\cite{xu2024distribution}. Additionally, a continual compatible training method: CVS~\cite{wan2022continual} is also introduced for a comprehensive comparison. Table~\ref{tab: setting1} and~\ref{tab: setting2} present the performance on Order-1 and Order-2, where C\textsuperscript{2}R\textsuperscript{*} represents the reproduced performance on the baseline method~\cite{xu2024lstkc} for fair comparison. The best results are marked as \textcolor{red}{\textbf{Red}}, and the second-best results are marked as \textcolor{blue}{\textbf{Blue}}.

\begin{table}[tbp]

\caption{AF performance on training Order-1.}
    
\renewcommand{\arraystretch}{1.1}
    \small
    \centering
    \setlength{\tabcolsep}{0.5mm}{
    \begin{tabular}{l|c|c|c}
    \hline
        \multicolumn{1}{c|}{\multirow{2}[1]{*}{Method}} & RFL-ReID & L-ReID & Average\\
        \cline{2-4}
        & AF(mAP/R1)↓& AF(mAP/R1)↓& AF(mAP/R1)↓\\
    \hline
        AKA~\cite{pu2021lifelong} & 26.4/35.0 &12.9/13.5 	&20.5/24.3 \\
        LwF~\cite{li2017learning} & 24.8/30.8 &12.4/14.0 	&18.6/22.4 \\
        CVS~\cite{wan2022continual} & 21.0/25.7 &17.7/19.5     &19.4/22.6 \\
        DKP~\cite{xu2024distribution} &21.7/22.4 &6.2/5.1 	    &13.9/13.7 \\
        PatchKD~\cite{sun2022patch} & 16.5/18.5 &7.6/7.1 	&12.1/12.8 \\
        LSTKC~\cite{xu2024lstkc} &11.3/11.6 &7.0/5.4 	&9.1/8.5 \\
        
        \hline
        \rowcolor{gray!15} C\textsuperscript{2}R &13.9/14.5 &7.8/7.8 	    &10.6/11.1 \\
        \rowcolor{gray!15} Bi-C\textsuperscript{2}R &\textbf{7.9}/\textbf{8.5} &\textbf{7.4}/\textbf{5.8} 	    &\textbf{7.6}/\textbf{7.1} \\
    \hline
    \end{tabular}
    }
    \label{tab: af}
\end{table}

\begin{figure*}[ht]
\centering
    \begin{minipage}[t]{\textwidth}
    \centering
    \includegraphics[width=0.22\textwidth]{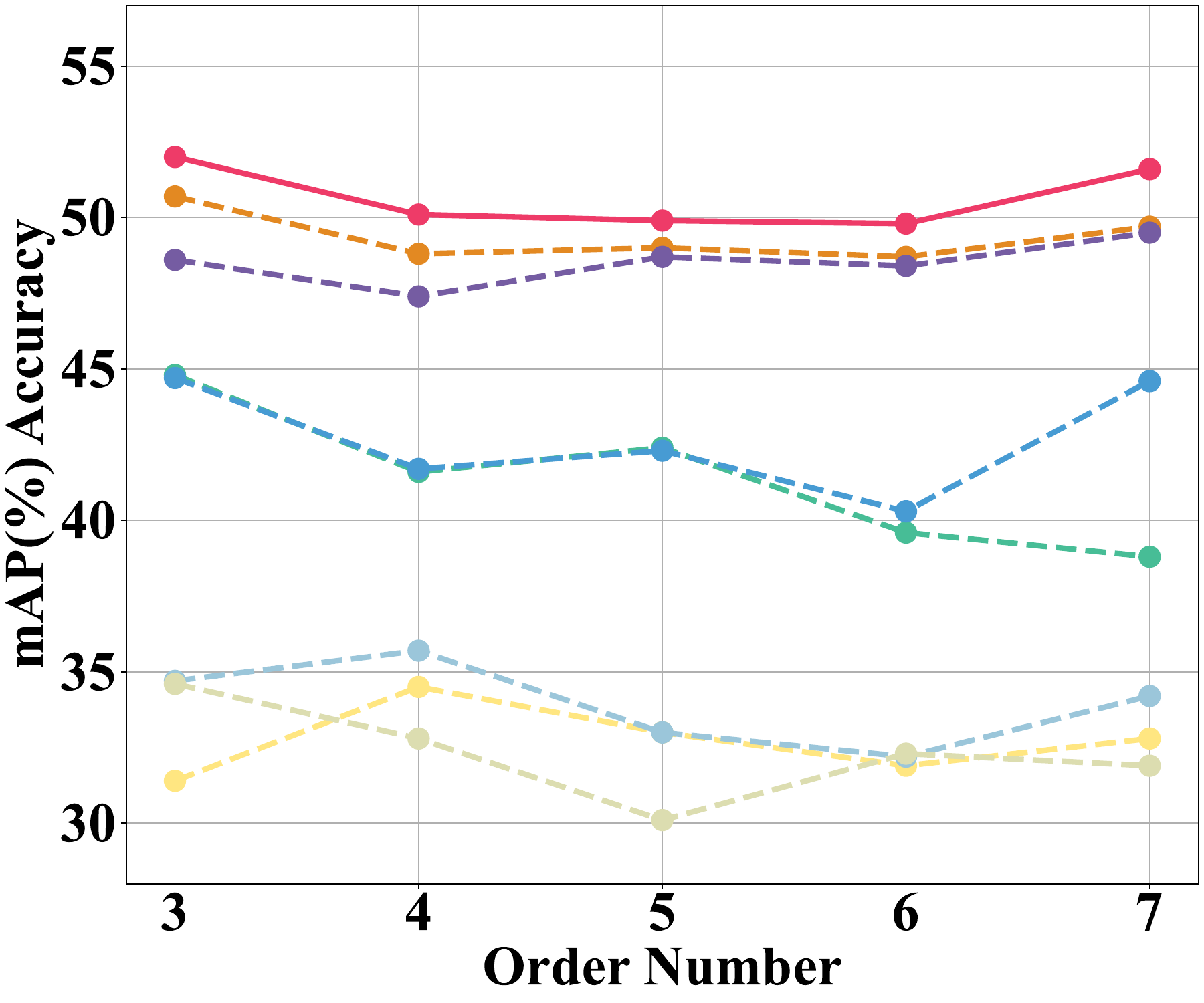}
    \includegraphics[width=0.22\textwidth]{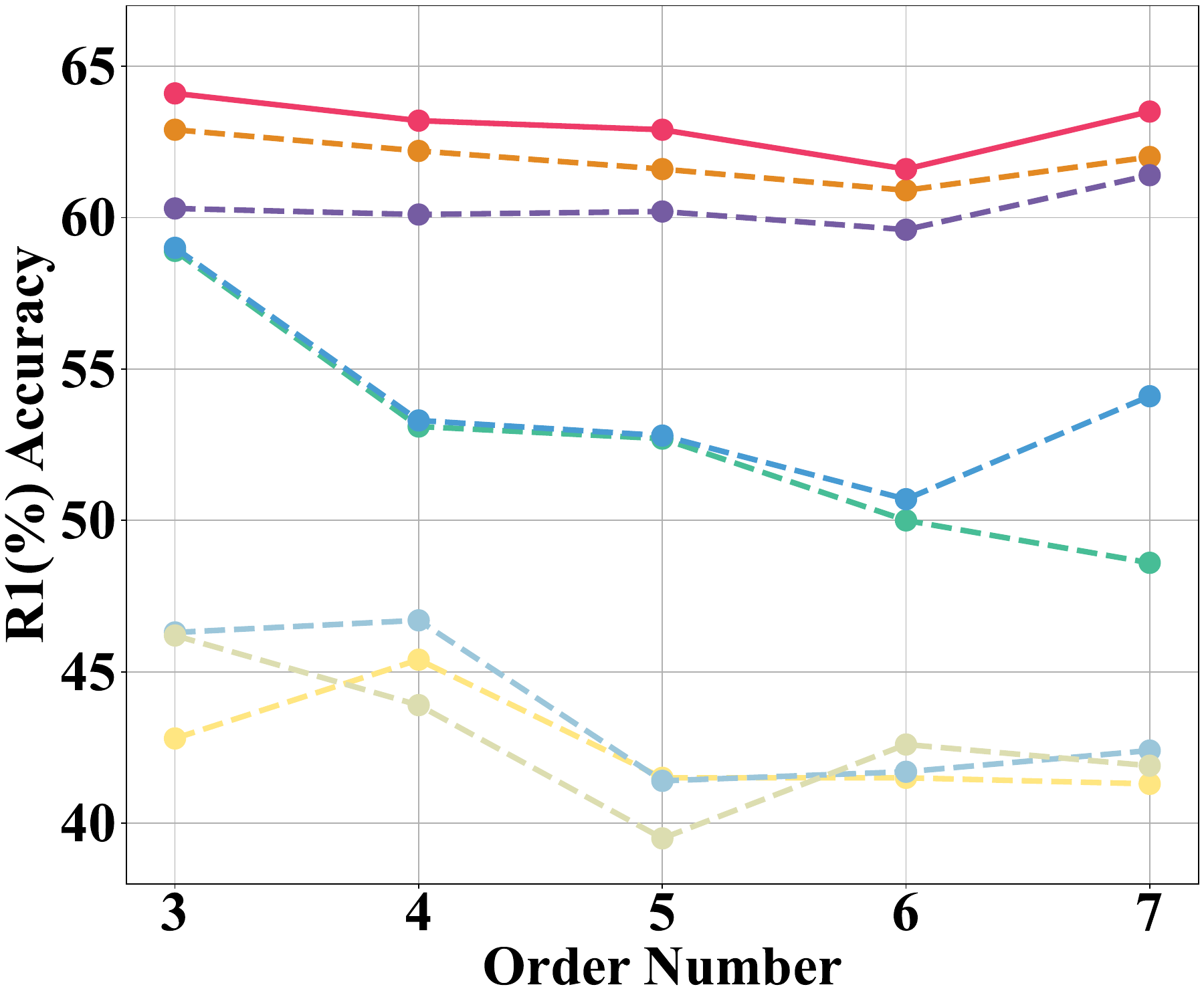}
    \includegraphics[width=0.22\textwidth]{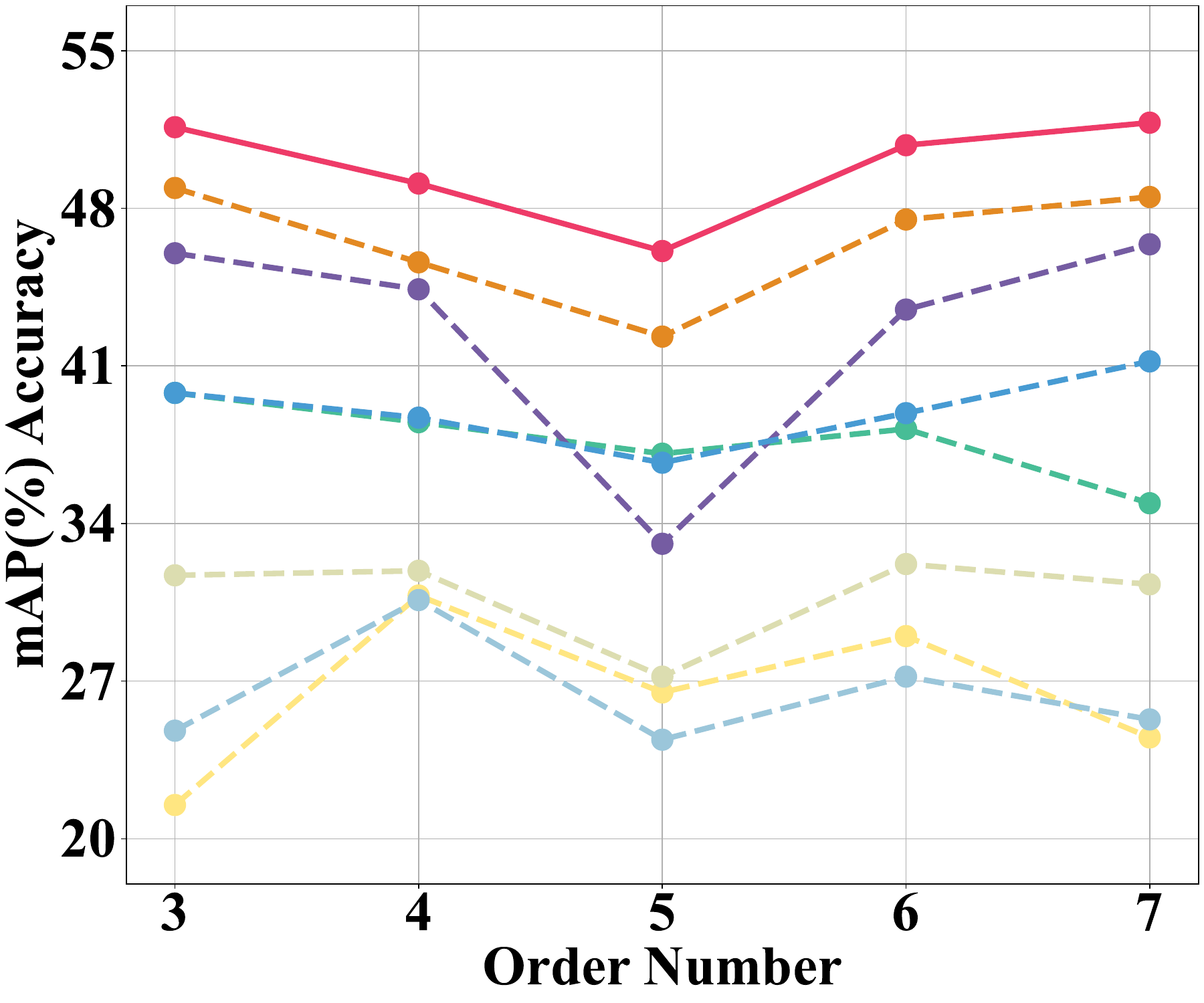}
    \includegraphics[width=0.22\textwidth]{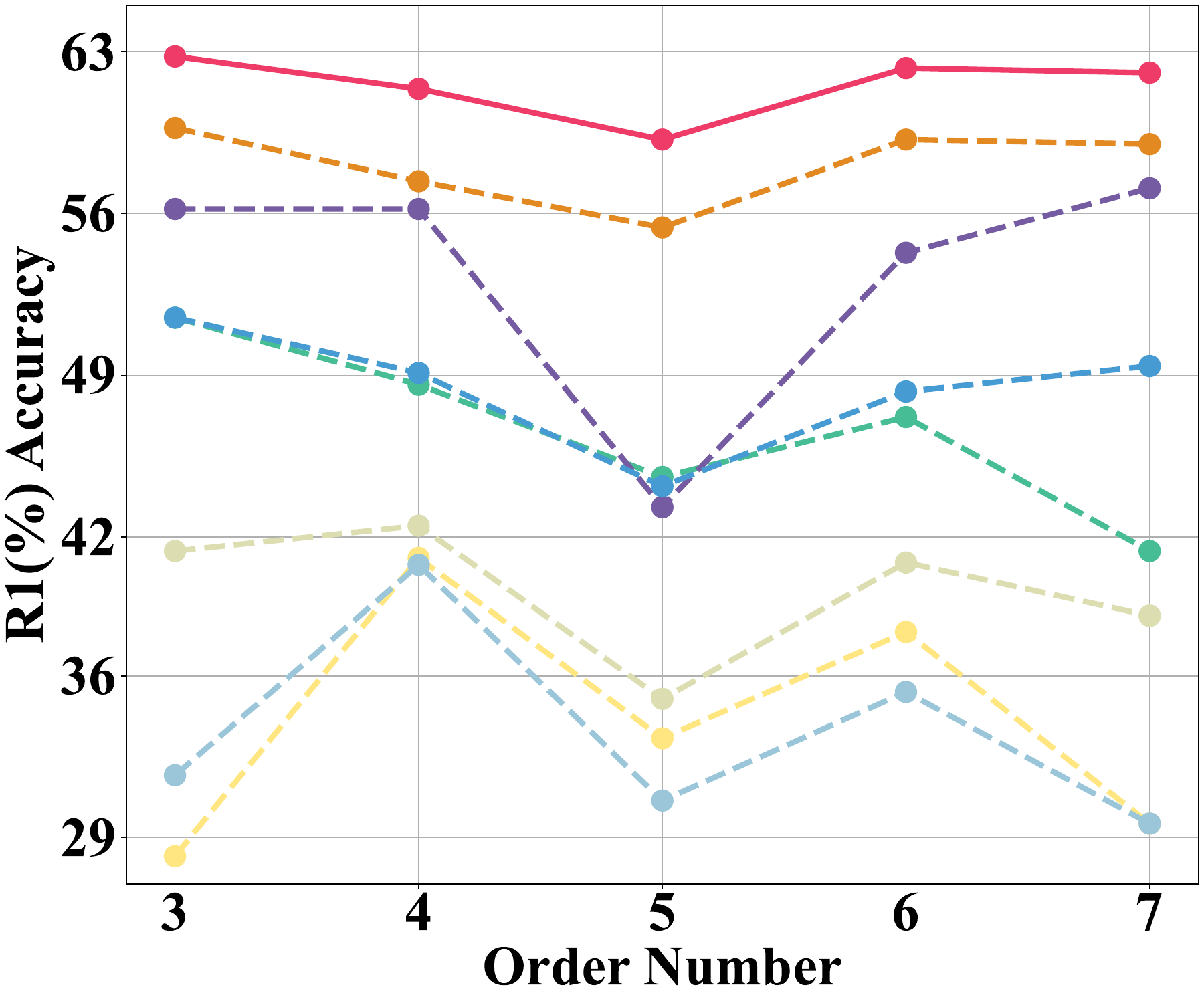}
    \includegraphics[width=0.075\textwidth]{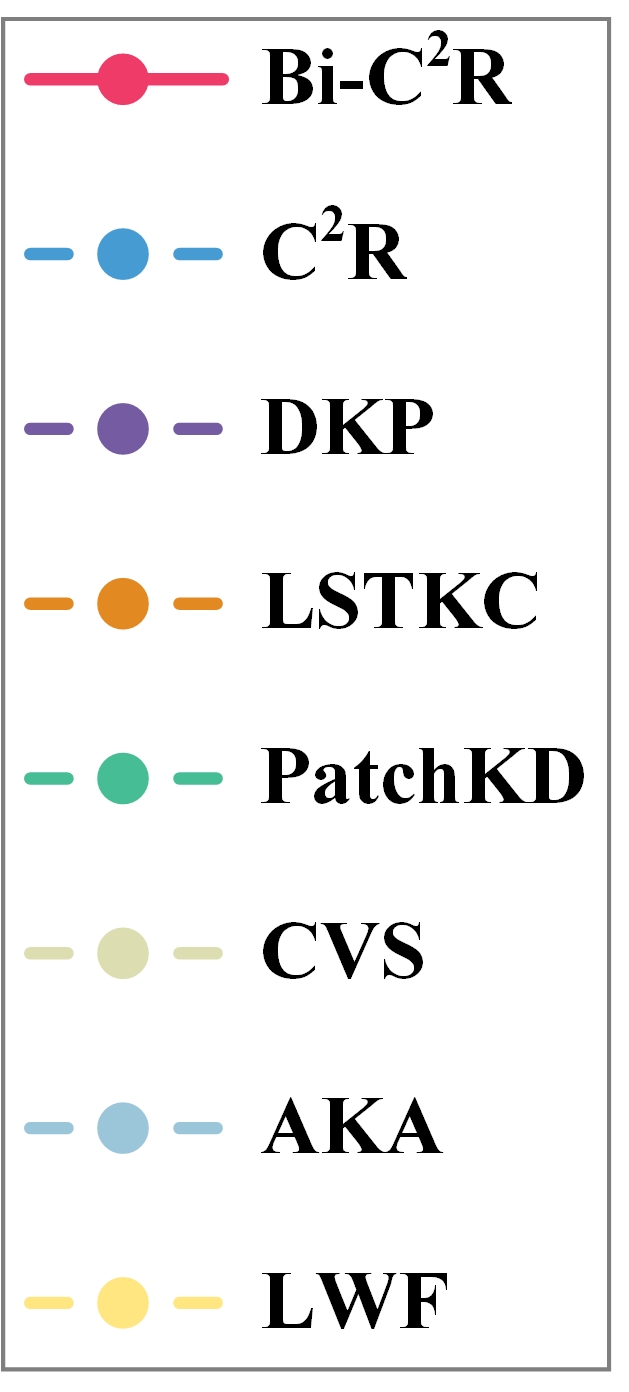}
    \end{minipage}
\caption{Performance on More Training Orders.}
\label{fig: orders}
\end{figure*}

\begin{figure*}[ht]
\centering
    \begin{minipage}[t]{\textwidth}
    \centering
    \includegraphics[width=0.22\textwidth]{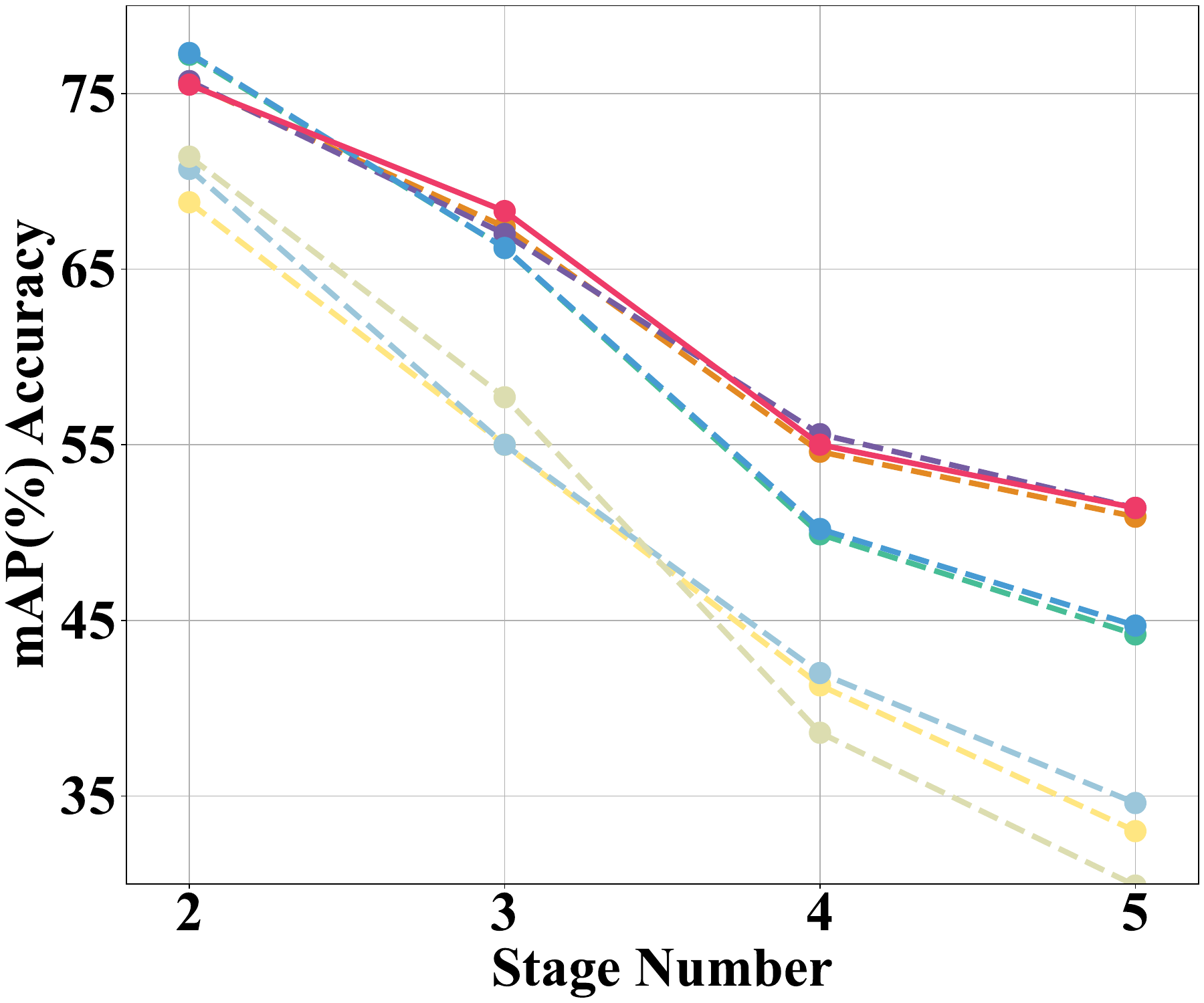}
    \includegraphics[width=0.22\textwidth]{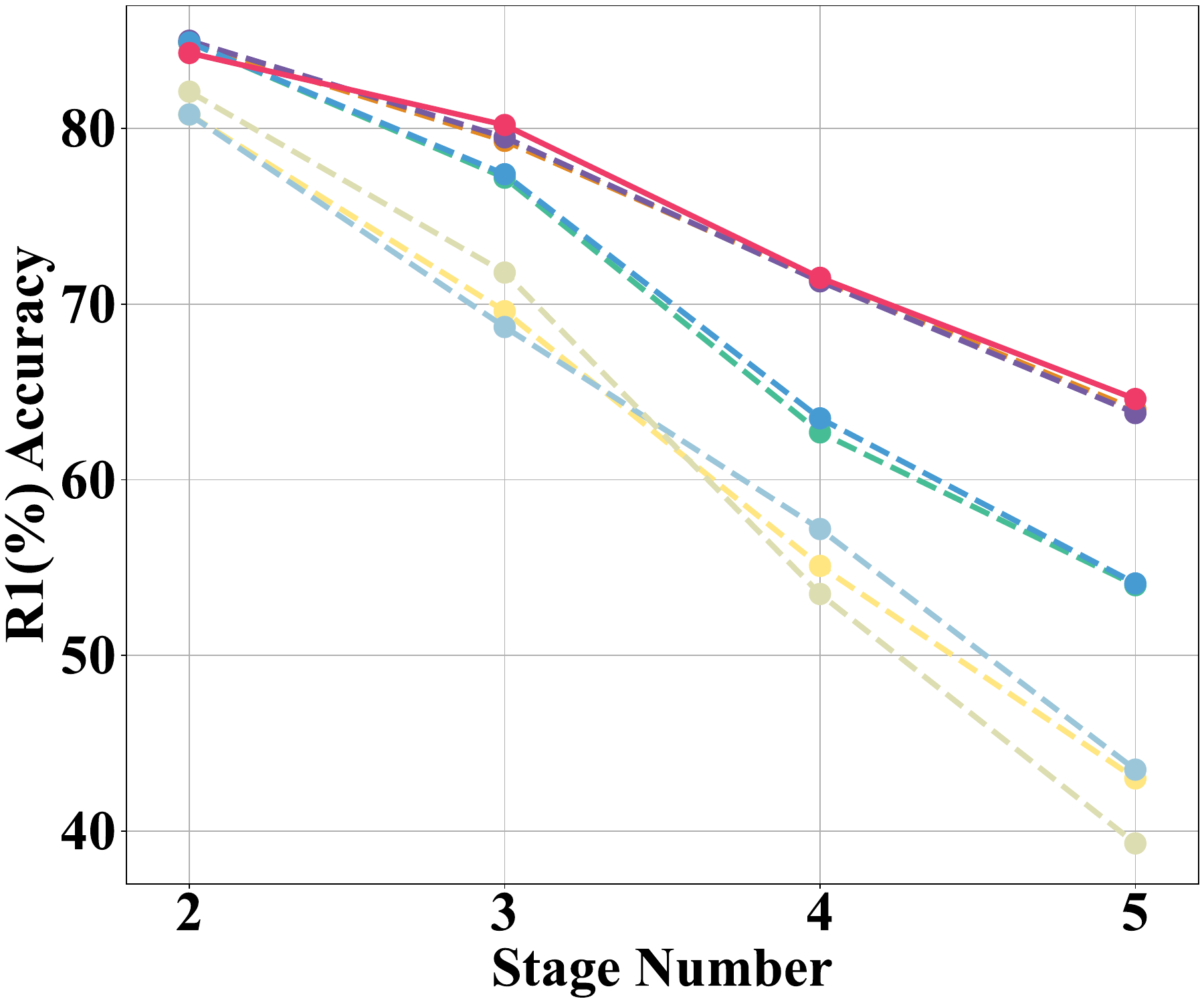}
    \includegraphics[width=0.22\textwidth]{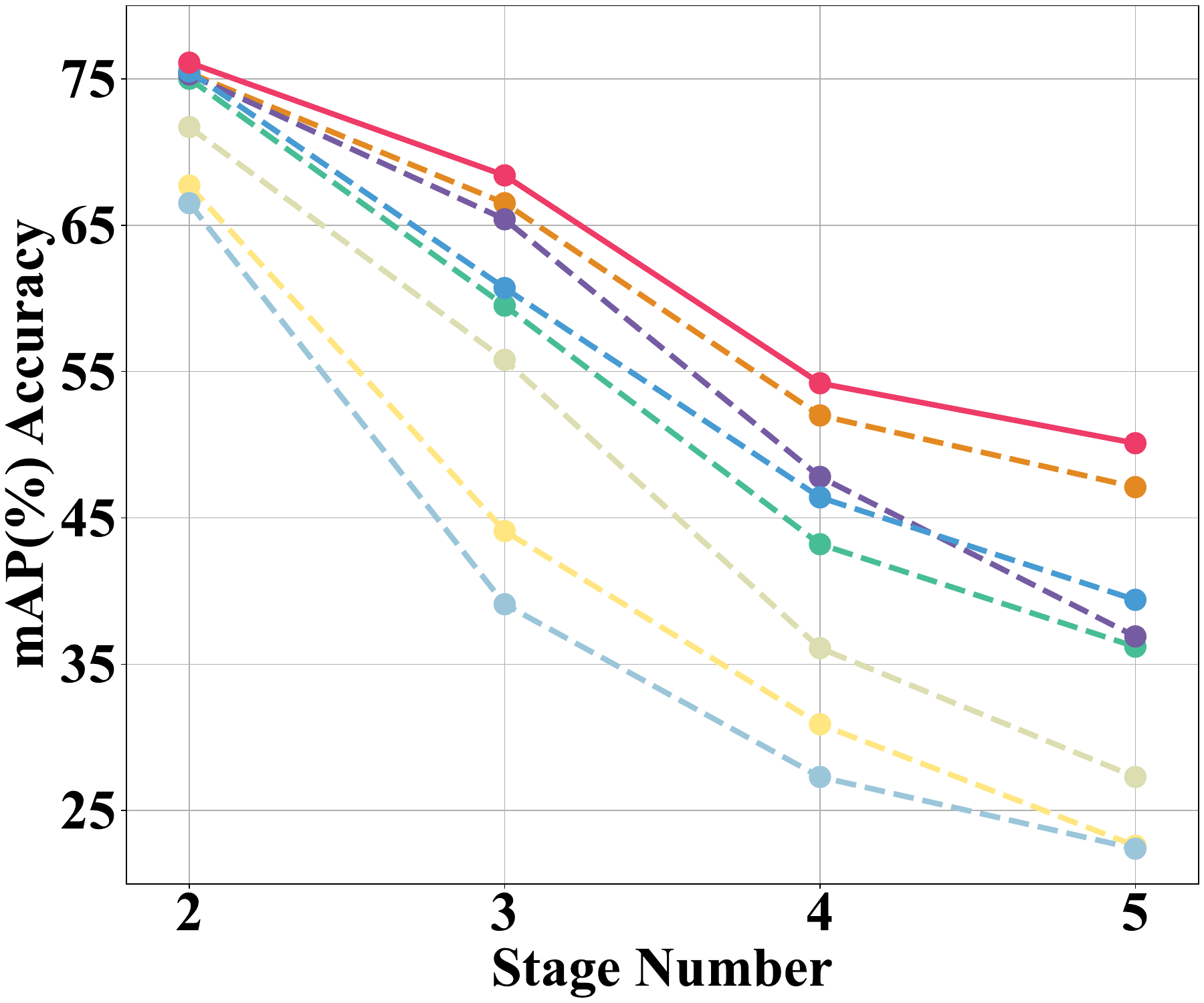}
    \includegraphics[width=0.22\textwidth]{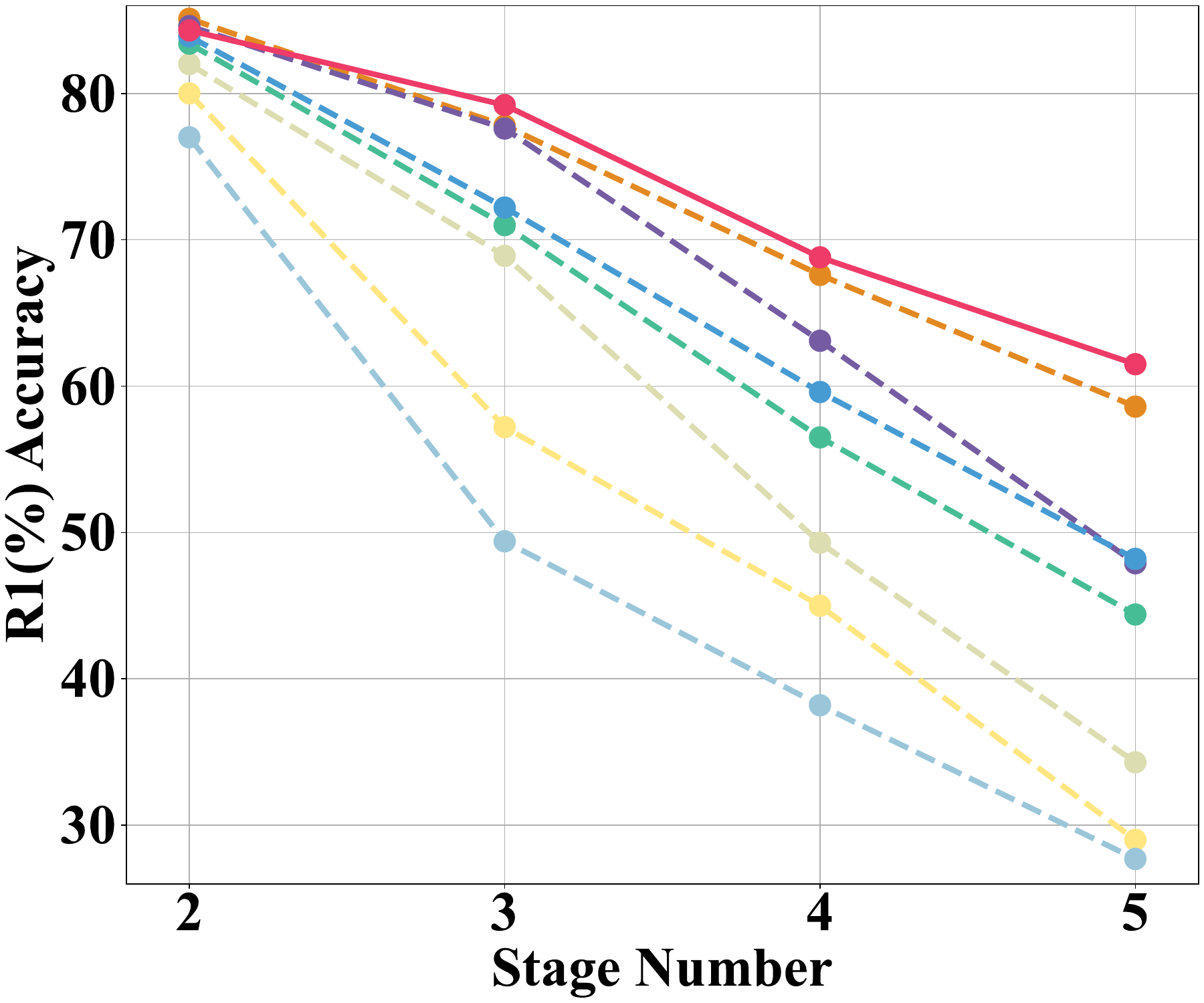}
    \includegraphics[width=0.075\textwidth]{fig/tuli.jpg}
    \end{minipage}
\caption{Performance of Different Stages on Training Order-1.}
\label{fig: stages}
\end{figure*}

\noindent\textbf{Comparison in the RFL-ReID task.} We first evaluate our Bi-C\textsuperscript{2}R by comparing it to other advanced L-ReID methods, which are reproduced by their officially released code. As shown in Table~\ref{tab: setting1} and~\ref{tab: setting2}, our Bi-C\textsuperscript{2}R achieves \textbf{50.1\%}/\textbf{61.5\%}(mAP/R1) on Order-1 and \textbf{51.2\%}/\textbf{63.2\%}(mAP/R1) on Order-2, which significantly outperforms the existing SOTA method LSTKC by \textbf{3.0\%}/\textbf{2.9\%} and \textbf{2.7\%}/\textbf{3.3\%}. This is because our method can continuously update old features in a compatible way, where old gallery features can be dynamically updated and anti-forgettable, thereby avoiding the incompatibility between query and gallery features without re-indexing old data effectively.

In particular, compared to the previous method C\textsuperscript{2}R, our extended Bi-C\textsuperscript{2}R method improves mAP and R1 accuracy by \textbf{2.5\%}/\textbf{2.1\%} on Order-1. The improvement of Bi-C\textsuperscript{2}R can be attributed to our bidirectional compatible transfer and dynamic feature fusion. Specifically, the bidirectional compatible transfer network pursues the mutual alignment of new and old knowledge, which facilitates the discriminability of new knowledge in the old feature space while updating the old features. Therefore, it enables the new model to obtain more common discriminative knowledge from both new and old data, further alleviating the forgetting of old knowledge. In addition, the dynamic feature fusion mitigates the severe feature drift problem derived from the EMA-based L-ReID model in the inference phase, thereby alleviating the incompatibility between the updated gallery features and query features.

\begin{table}[tbp]
\caption{AF performance of Market1501 on training Order-1.}
\renewcommand{\arraystretch}{1.1}
    \small
    \centering
    \setlength{\tabcolsep}{0.8mm}{
    \begin{tabular}{c|l|c|c|c|c|c|c|c|c}
    \hline
        \multicolumn{1}{c|}{\multirow{2}[1]{*}{Task}} &\multicolumn{1}{c|}{\multirow{2}[1]{*}{Method}} & \multicolumn{2}{c|}{Stage 2} & \multicolumn{2}{c|}{Stage 3} & \multicolumn{2}{c|}{Stage 4} & \multicolumn{2}{c}{Stage 5} \\
        \cline{3-10}
        & & mAP & R1 & mAP & R1 & mAP & R1 & mAP & R1\\
        \hline
        \multirow{4}[1]{*}{\rotatebox[origin=c]{90}{RFL-ReID}} & DKP & 2.0 & 0.6 & 8.4 & 5.1 & 22.1 & 18.0 & 37.3 & 35.8 \\
        & LSTKC & 4.4 & 2.5 & 8.9 & 6.7 & 19.6 & 17.1 & 23.4 & 22.5 \\
        & PatchKD & 2.4 & 2.7 & 11.0 & 8.8 & 16.1 & 12.5 & 16.1 & 12.7 \\
        & \cellcolor{gray!15}Bi-C\textsuperscript{2}R & \cellcolor{gray!15}1.5 & \cellcolor{gray!15}2.3 & \cellcolor{gray!15}5.6 & \cellcolor{gray!15}5.8 & \cellcolor{gray!15}13.5 & \cellcolor{gray!15}10.0 & \cellcolor{gray!15}15.7 & \cellcolor{gray!15}11.7 \\
        \hline
        \multirow{4}[1]{*}{\rotatebox[origin=c]{90}{L-ReID}} & LSTKC & 3.9 & 2.9 & 9.3 & 6.1 & 17.5 & 10.9 & 17.6 & 10.9 \\
        & DKP & 1.2 & 0.2 & 7.7 & 3.8 & 11.8 & 5.9 & 14.0 & 8.2 \\
        & PatchKD & 1.5 & 0.2 & 3.7 & 1.6 & 9.1 & 4.5 & 9.9 & 5.1 \\
        & \cellcolor{gray!15}Bi-C\textsuperscript{2}R & \cellcolor{gray!15}0.2 & \cellcolor{gray!15}0.1 & \cellcolor{gray!15}3.6 & \cellcolor{gray!15}1.6 & \cellcolor{gray!15}8.5 & \cellcolor{gray!15}4.3 & \cellcolor{gray!15}9.6 & \cellcolor{gray!15}4.7 \\
    \hline
    \end{tabular}
    }
    \label{tab: af_old}
\end{table}

\noindent\textbf{Comparison on the traditional L-ReID task.} Rather than the evaluation in the RFL-ReID task, we further evaluate our Bi-C\textsuperscript{2}R on the traditional L-ReID task, as shown in Table~\ref{tab: setting1} and~\ref{tab: setting2}. The experimental results show that our Bi-C\textsuperscript{2}R achieves advanced performance to the existing SOTA method DKP on both orders, which achieves \textbf{51.4\%}/\textbf{64.6\%}(mAP/R1) and \textbf{51.6\%}/\textbf{64.4\%}(mAP/R1), respectively. The comparable performance on traditional L-ReID implies that our bidirectional compatible transfer method does not sacrifice the discriminability of the new model. Therefore, our method can also be applied to the general scenario when old gallery data are available.

Benefiting from the above design, our Bi-C\textsuperscript{2}R preserves more common discriminative knowledge by mutually mapping new and old feature spaces to each other, which improves the unified discriminability of new and old knowledge by bi-directional feature transfer and dynamic feature fusion in both training and inference phases.

\begin{table}[tbp]
\caption{Ablation study on fixed balancing factors in BiCT network.}
\renewcommand{\arraystretch}{1.1}
\small
    \centering
    \begin{tabular}{c|c|c|c|c|c|c|c}
    \hline
        $a^m$ & 0.0 & 0.2 & 0.4 & 0.6 & 0.8 & 1.0 & \textbf{Ours}\\
    \hline
        mAP & 49.3 & 49.6 & 49.7 & 49.8 & 49.6 & 49.7 & \textbf{50.1}\\
    \hline
        R1 & 60.8 & 61.1 & 61.1 & 61.2 & 61.2 & 60.9 & \textbf{61.5}\\
    \hline
    \end{tabular}
    \label{tab: ablation_cct}
\end{table}

\begin{table}[tbp]
\caption{Ablation study on component modules in BiCT network.}
\renewcommand{\arraystretch}{1.1}
\small
    \centering
    \begin{tabular}{cc|c|c|c|c|c|c}
    \hline
        \multirow{2}[1]{*}{FMM} & \multirow{2}[1]{*}{KCM} & \multicolumn{2}{c|}{Market1501} & \multicolumn{2}{c|}{CUHK03} & \multicolumn{2}{c}{Average}\\
    \cline{3-8}
        & & mAP & R1 & mAP & R1 & mAP & R1\\
    \hline
        \checkmark &  & 57.2 & 71.7 & 41.7 & 42.2 & 49.6 & 61.2 \\
    \hline
        & \checkmark & 55.8 & 70.5 & 43.5 & 44.6 & 49.9 & 61.0\\
    \hline
        \checkmark & \checkmark & 57.1 & 71.2 & 43.6 & 44.5 & \textbf{50.1} & \textbf{61.5}\\
    \hline
    \end{tabular}
    \label{tab: ablation_cct_module}
\end{table}

\noindent\textbf{Comparison under more practical scenarios.} In addition to the widely adopted benchmarking orders mentioned above, we further evaluated our Bi-C\textsuperscript{2}R across different camera configurations, lighting conditions, and clothing variations in real-world deployment scenarios. Therefore, we introduced two practical and challenging datasets (\emph{i.e.}, LTCC and PRCC) with greater differences in the clothing, lighting, and camera in Order-8 and Order-9 to evaluate the robustness of our method. As reported in Tables 1 and 2, our method significantly outperforms one of the SOTA methods, LSTKC, on both orders on average mAP/R1. These results can be attributed to our proposed continual compatible feature updation, which adaptively balances diverse data from different domains, achieving more compact intra-class representations regardless of changes in people's clothing, lighting conditions, and imaging cameras. Therefore, the above results further validate the robustness of our proposed Bi-C\textsuperscript{2}R.

\subsection{Comparison of Average Forgetting Performance}
In addition to the above ReID performance comparison, we further report the AF performance of mAP and R1 on Order-1 to evaluate the anti-forgetting performance in Table~\ref{tab: af}. The experimental results show that our Bi-C\textsuperscript{2}R realizes the lowest average forgetting performance, which achieves an average of \textbf{7.6\%}/\textbf{7.1\%} on the mAP/R1 for both RFL-ReID and L-ReID tasks, leading our previous method C\textsuperscript{2}R by \textbf{3.0\%}/\textbf{4.0\%}. The anti-forgetting ability of our BiC2R is three-folded. Firstly, our forward transfer network achieves compatibility between new and old features by compatible feature distilling and anti-forgetting of old knowledge distilling. Secondly, the backward transfer network facilitates the common discrimination between both new and old knowledge by reconstructing old discriminative knowledge based on new features. Finally, the dynamic feature fusion module further aligns the transferred features with the new model after parameter fusion in the inference phase. As a result, our Bi-C\textsuperscript{2}R achieves the highest anti-forgetting performance by continuously updating old features after each training stage.

\begin{figure}[tbp]
\vspace{-14pt}
\centering
\subfloat[Total Time on Order-1]{\includegraphics[width=0.24\textwidth]{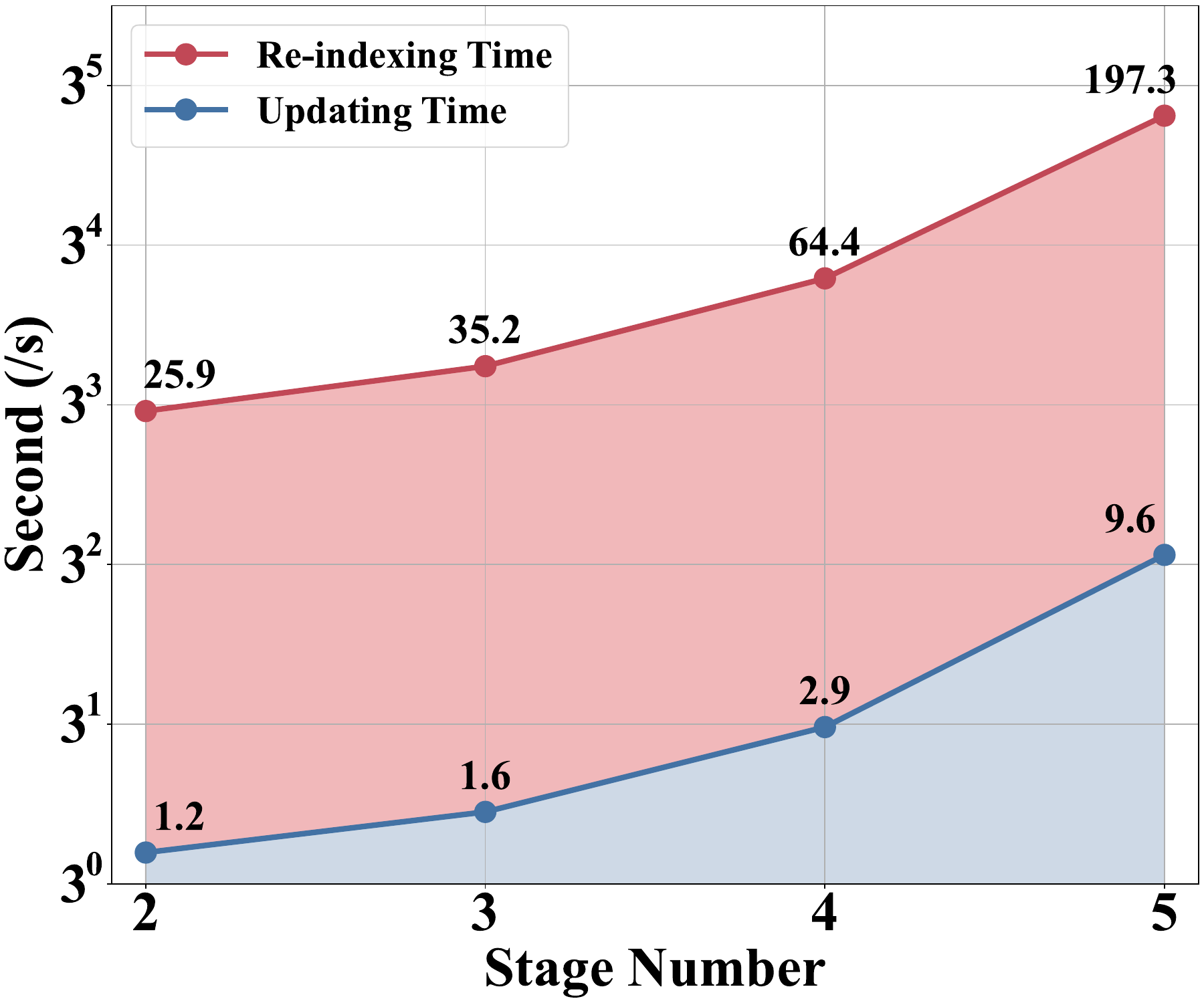}}
\subfloat[Batch Time on Order-1]{\includegraphics[width=0.24\textwidth]{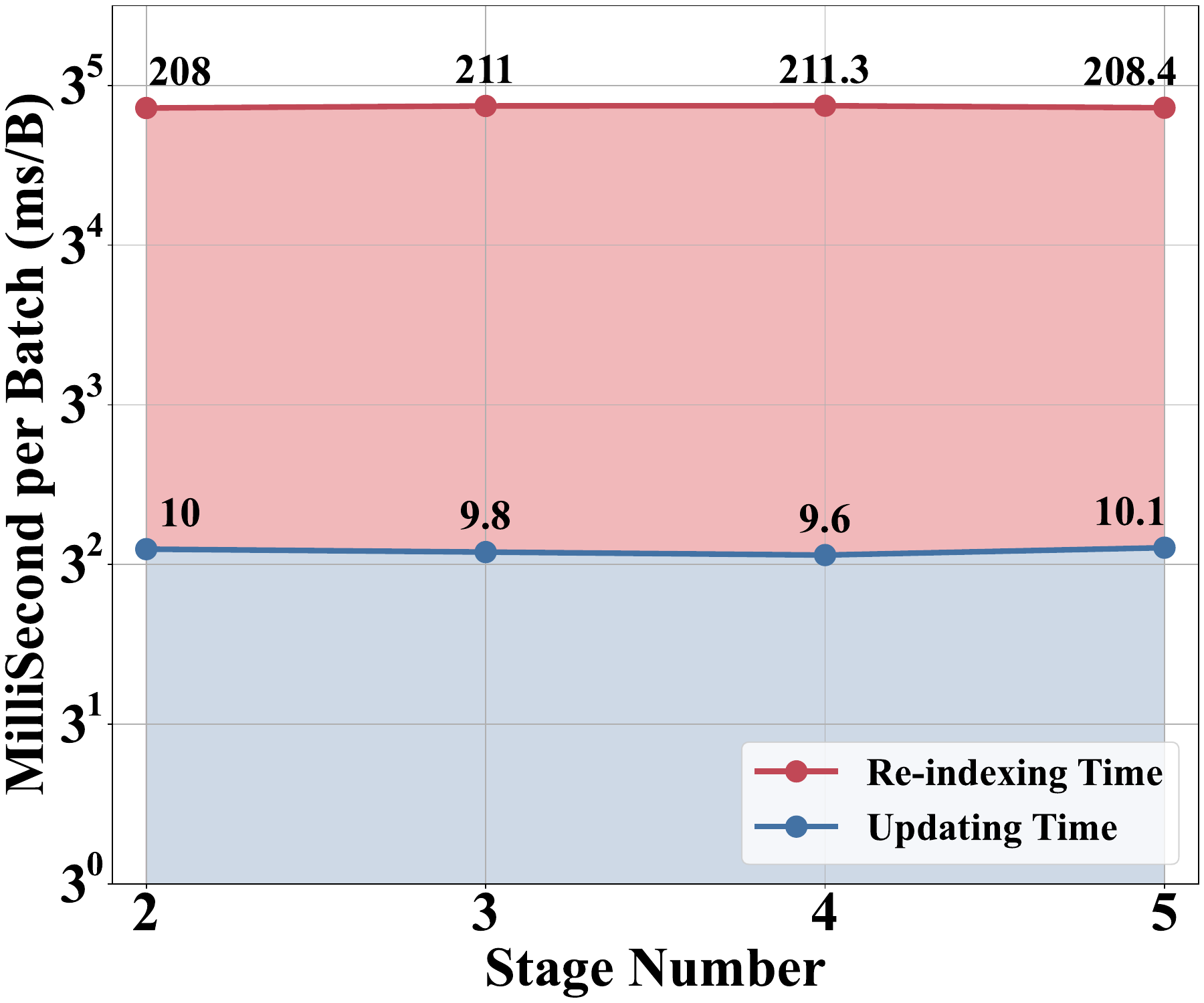}}

\subfloat[Total Time on Order-2]{\includegraphics[width=0.24\textwidth]{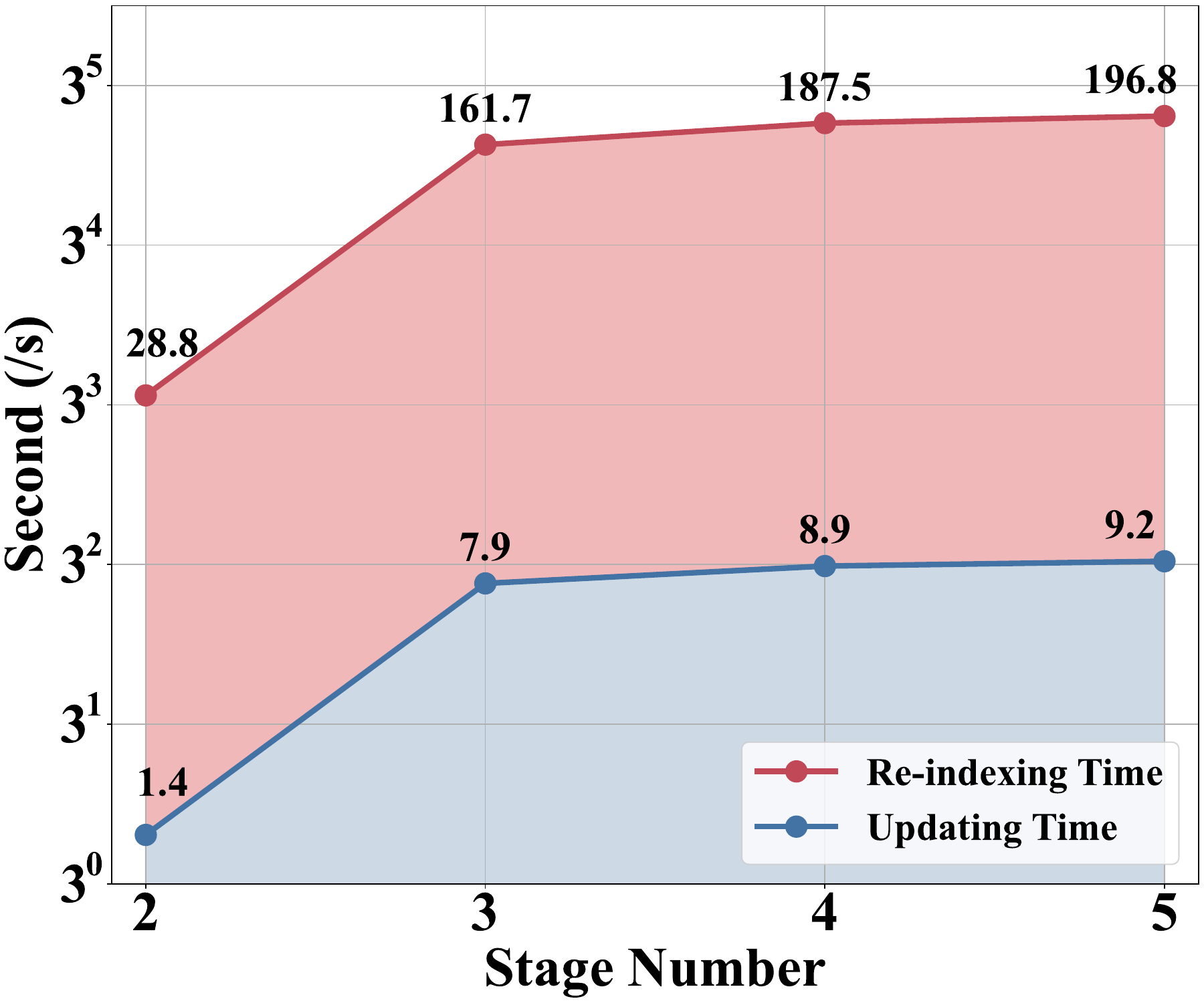}}
\subfloat[Batch Time on Order-2]{\includegraphics[width=0.24\textwidth]{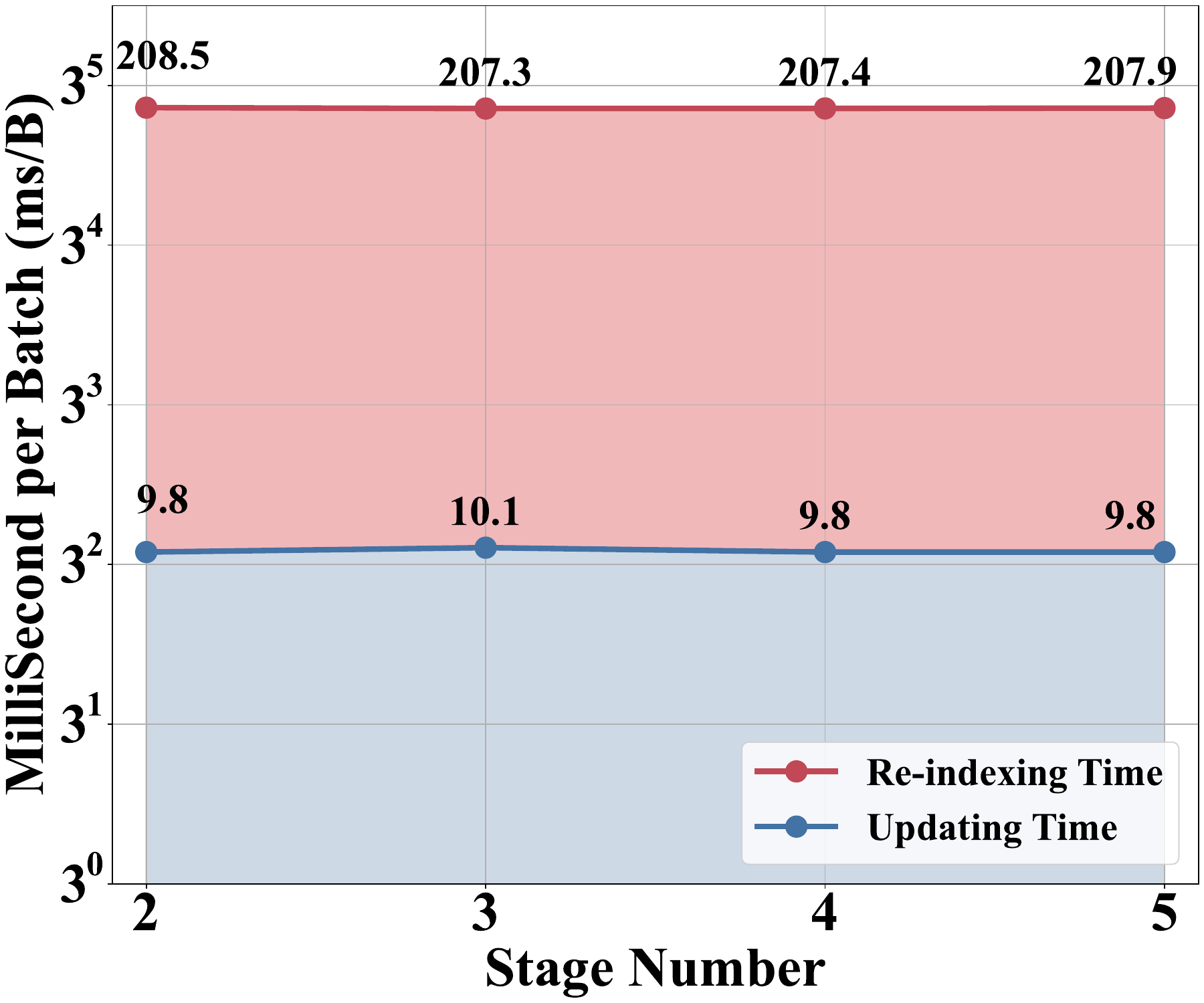}}
\caption{The comparison of time cost between the feature updating strategy and the feature re-indexing strategy in our Bi-C\textsuperscript{2}R.}
\label{fig: efficiency}
\end{figure}

\begin{table}[tbp]
\caption{Ablation on choices of feature fusion strategy.}
\renewcommand{\arraystretch}{1.1}
\small
    \centering
    \begin{tabular}{c|c|c}
    \hline
        Strategy & mAP & R1\\
    \hline
        w/o Fusion & 49.0 & 60.9 \\
    \hline
        Fixed (0.5) & 49.7 & 61.2 \\
        Time-increasing & 49.2 & 60.9 \\
        Time-descending & 49.6 & 61.0 \\
    \hline
        DFF (Ourss) & 50.1 & 61.5 \\
    \hline
    \end{tabular}
    \label{tab: ablation_fuse}
\end{table}

\par Furthermore, to intuitively demonstrate the old knowledge preserving capacity of our Bi-C\textsuperscript{2}R, we present the forgetting rates of different models for old knowledge (AF on Market-1501) in Table~\ref{tab: af_old}, where a lower forgetting rate represents greater memory retention after new identities are continuously collected. It can be seen that our Bi-C\textsuperscript{2}R significantly leads DKP and LSTKC after continuously training at all stages. Although PatchKD also achieves compatible preserving performance, it significantly sacrifices the capacity to learn new knowledge (performance on CUHK03 in Table~\ref{tab: setting1}). The above results can be attributed to our BCD module, which achieves compatibility between old and new features by aligning transferred features with new features (as in Eq.~\ref{eq:l_ca}), thereby suppressing forgetting caused by changes of the distribution between old and new knowledge. Furthermore, by preserving the old knowledge space composed of affinities between different individuals (as in Eq.~\ref{eq:bcd_kl}), the BCD module explicitly preserves the relative positions of old knowledge after the feature transfer, thus achieving the optimal trade-off between new and old knowledge.

\begin{table}[t]
\caption{Ablation on alternative methods in BiCT network.}
\renewcommand{\arraystretch}{1.1}
\small
    \centering
    \begin{tabular}{c|c|c|c|c}
    \hline
        Methods & Full-Conv & Full-Linear & Full-Trans & \textbf{BiCT-Net}\\
    \hline
        mAP & 48.9 & 49.4 & 49.2 & \textbf{50.1}\\
    \hline
        R1 & 59.9 & 60.9 & 60.6 & \textbf{61.5}\\
    \hline
    \end{tabular}
    \label{tab: ablation_cct_alter}
\end{table}

\subsection{Comparison of Inference Efficiency}
\par To further verify the inference efficiency of our proposed method, we also present the comparison of the inference efficiency, including the total time (as shown in Figure~\ref{fig: efficiency} (a) and (c)) and the batch time (as shown in Figure~\ref{fig: efficiency} (b) and (d)) at different stages on Order-1 and Order-2, where the batch size is set as 128. It can be seen that the feature updating strategy in our Bi-C\textsuperscript{2}R achieves an acceleration of over $\times$20 at each training stage, which also achieves a similar advantage in batch time. This improvement comes from our lightweight forward transfer network, which not only does not rely on old gallery images but also enables fast and effective feature transfer of old gallery features by balancing new knowledge acquisition and old knowledge preservation. Therefore, the above results further verify the superiority of our Bi-C\textsuperscript{2}R in inference efficiency.

\begin{figure*}[ht]
\centering
\includegraphics[width=0.95\textwidth]{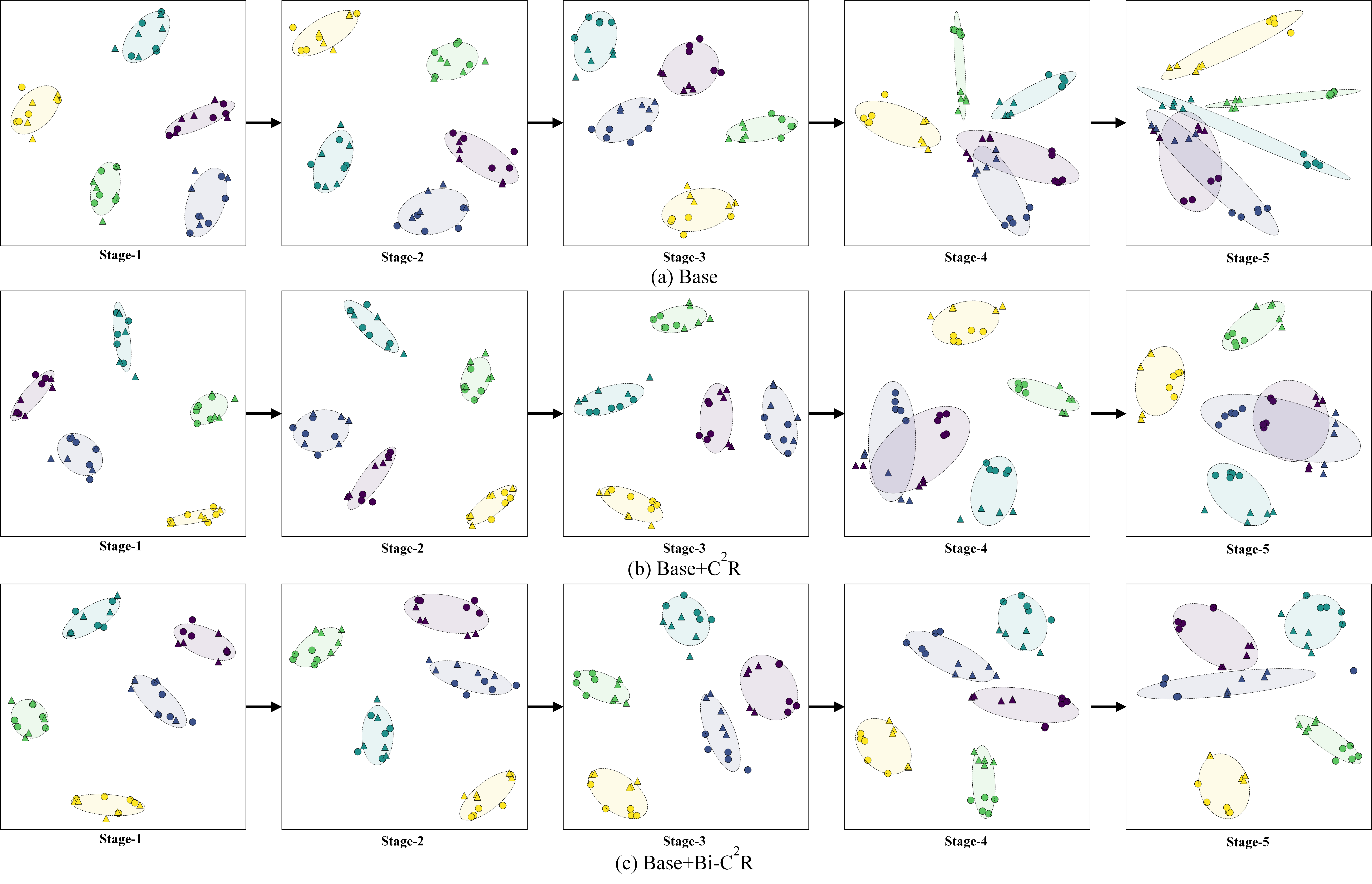}
\caption{The t-SNE visualization of different training stages of Market1501 dataset on Order-1, where different colours represent different identities and $\bigcirc, \triangle$ represent features of randomly sampled gallery images and query images.}
\label{fig: tsne}
\end{figure*}

\begin{figure*}[ht]
\centering
\includegraphics[width=0.95\textwidth]{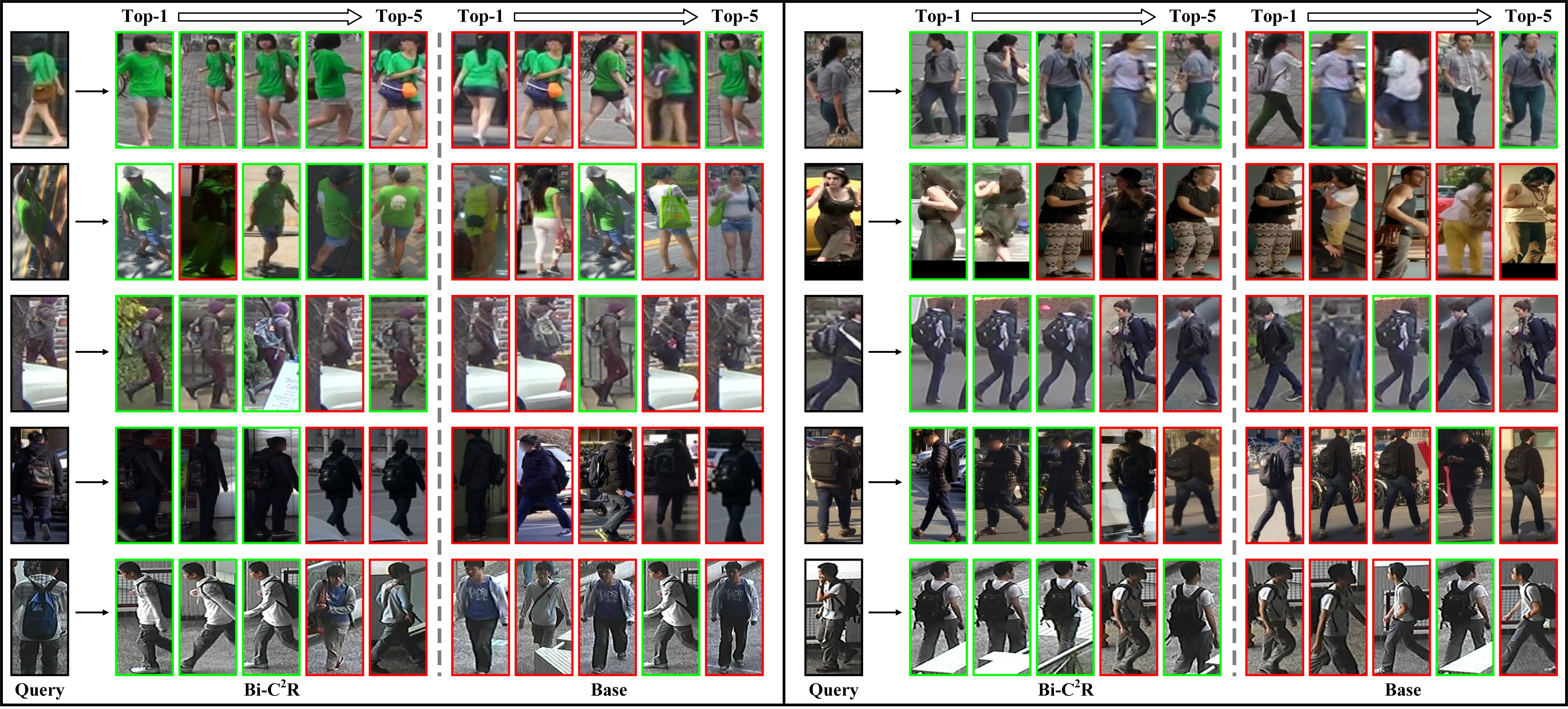}
\caption{Re-identification results of different datasets on Order-1.}
\label{fig: reid}
\end{figure*}

\subsection{Ablation Studies}
In this section, we conduct extensive ablation studies on Order-1 to evaluate the effectiveness of each component in our Bi-C\textsuperscript{2}R.

\begin{table}[tbp]

\caption{Ablation study about the influence of each component in Bi-C\textsuperscript{2}R.}

\renewcommand{\arraystretch}{1.1}
\small
    \centering    
    \setlength{\tabcolsep}{0.6mm}{
    \begin{tabular}{c|ccccc|c|c|c|c}
    \hline
        \multirow{2}[1]{*}{Method} & \multirow{2}[1]{*}{LSTKC\textsuperscript{\cite{xu2024lstkc}}} & \multirow{2}[1]{*}{BiCT} & \multirow{2}[1]{*}{BiCD} & \multirow{2}[1]{*}{BiAD} & \multirow{2}[1]{*}{DFF} & \multicolumn{2}{c|}{L-ReID} & \multicolumn{2}{c}{RFL-ReID}\\
    \cline{7-10}
        & & & & & & mAP & R1 & mAP & R1\\
        \hline
        Base & \checkmark &  &  &  &  & 50.0 & 63.1 & 47.1 & 58.6\\
        & \checkmark & \checkmark & \checkmark &  &  & 51.0 & 64.0 & 48.3 & 59.7\\
        & \checkmark & \checkmark & \checkmark & \checkmark &  & 51.4 & 64.6 & 49.0 & 60.9\\
        Bi-C\textsuperscript{2}R & \checkmark & \checkmark & \checkmark & \checkmark & \checkmark & 51.4 & 64.6 & 50.1 & 61.5\\
    \hline
    \end{tabular}
    }
    \label{tab: ablation}
\end{table}

\begin{figure}[htbp]
\vspace{-14pt}
\centering
\subfloat[The weight for $\mu_1$]{\includegraphics[width=0.24\textwidth]{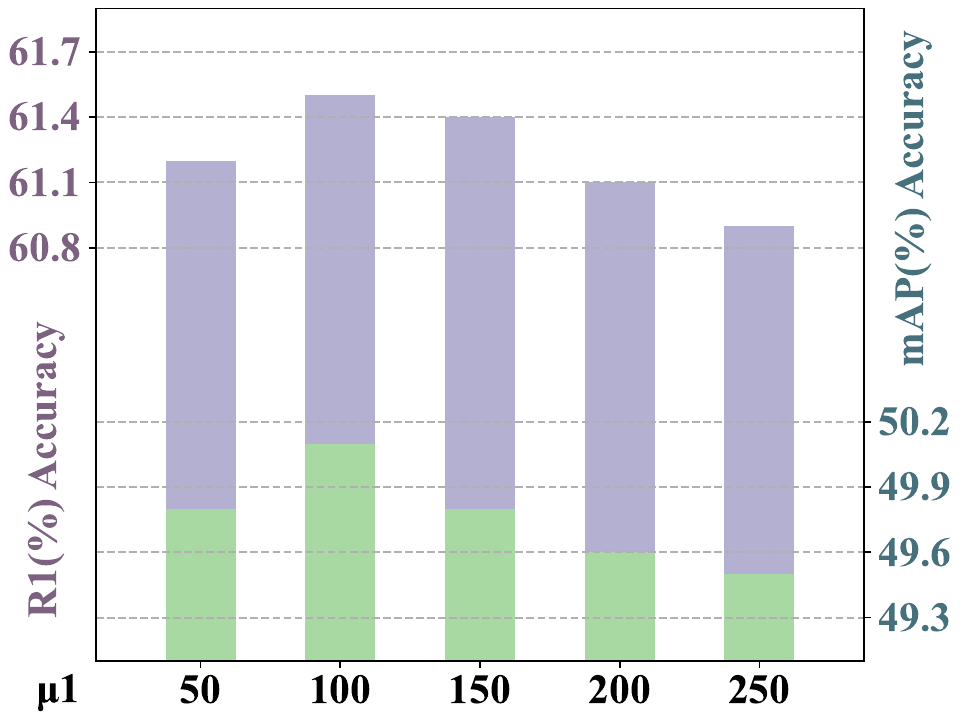}}
\subfloat[The weight for $\mu_2$]{\includegraphics[width=0.24\textwidth]{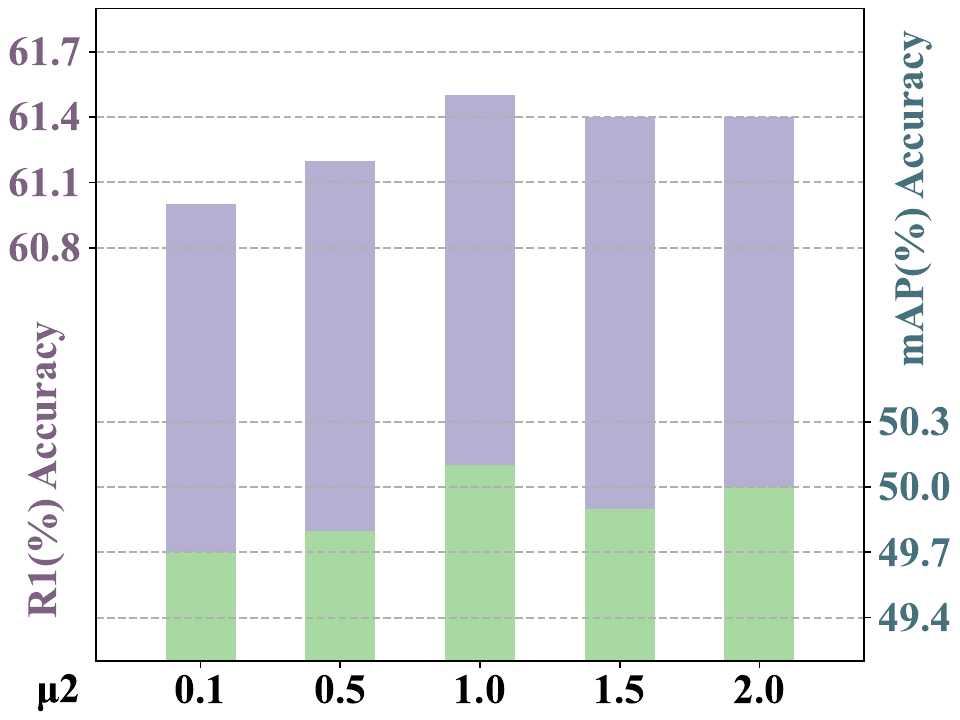}}

\subfloat[The weight for $\mu_3$]{\includegraphics[width=0.24\textwidth]{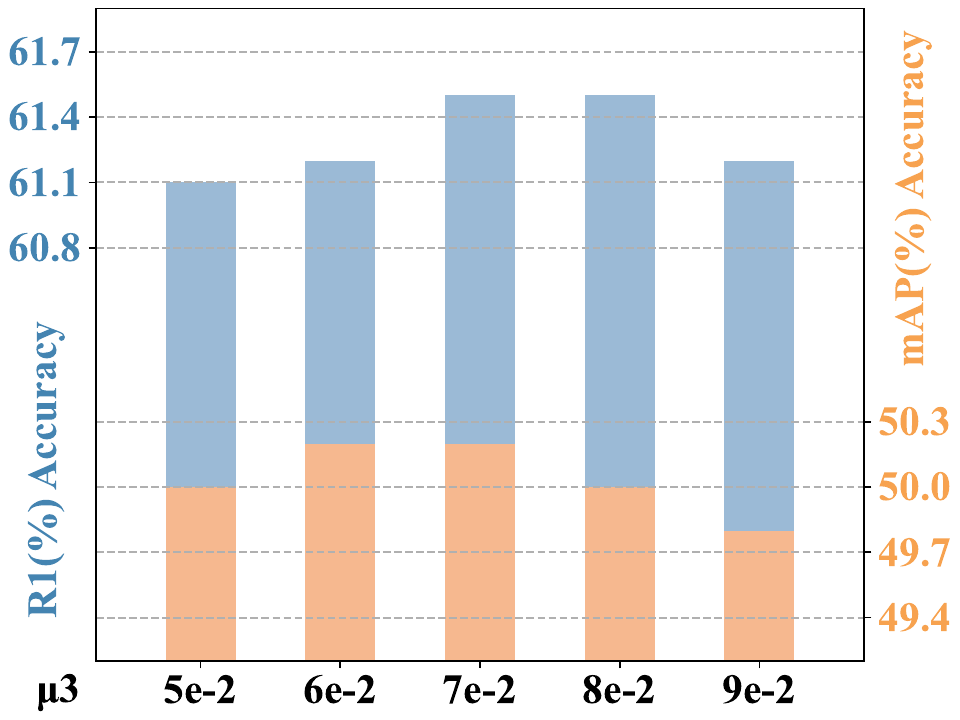}}
\subfloat[The weight for $\mu_4$]{\includegraphics[width=0.24\textwidth]{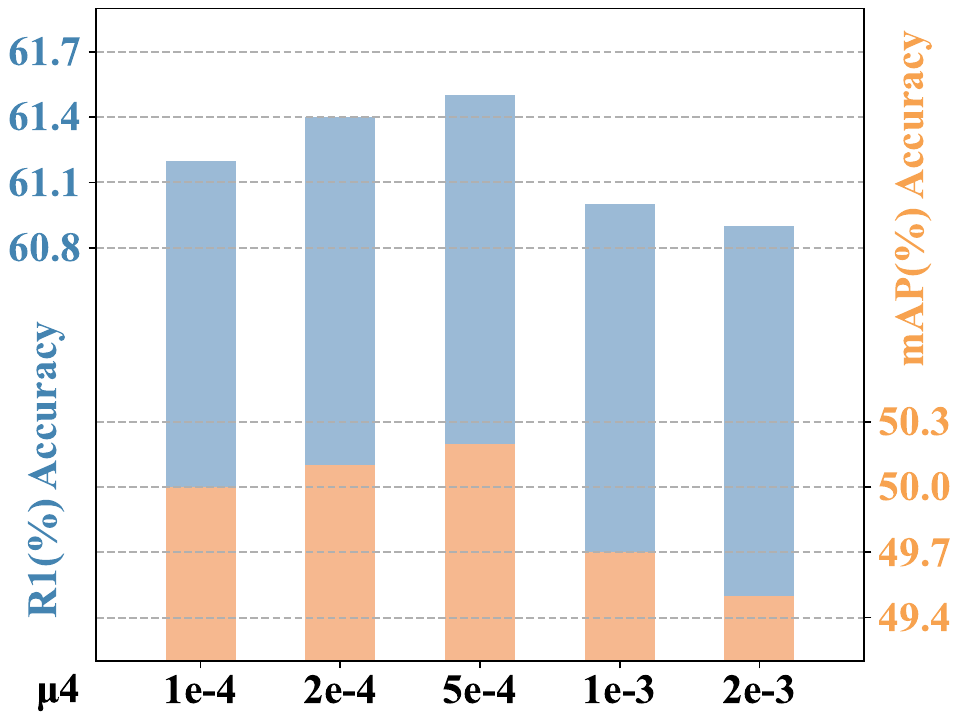}}
\caption{The influence of hyper-parameters in our Bi-C\textsuperscript{2}R.}
\label{fig: params}
\end{figure}

\par \noindent\textbf{Effectiveness of BiCT network.} To begin with, we verify the effectiveness of our BiCT network by comparing our adaptively balancing strategy to the fixed balancing strategy for $a^m$ in~\ref{eq:cct}. As shown in Table~\ref{tab: ablation_cct}, when the proportion of old knowledge increases, the average performance continues to grow, while outstanding performance cannot be achieved by setting a fixed weight of old knowledge due to the diverse discrepancy between different data domains. Instead, our method achieves the highest performance on both mAP and R1 (reaching \textbf{50.1\%}/\textbf{61.5\%}) by adaptively adjusting the proportion of new and old knowledge between different domains, verifying its superior robustness and compatibility.


\par In addition, as shown in Table~\ref{tab: ablation_cct_module}, we further evaluated the impact of two key components of BiCT on the retention of both old and new knowledge: the Forward Mapping Module (FMM) and the Knowledge Capturing Module (KCM). It can be seen that when only the FMM is employed, the system tends to retain old knowledge, achieving the highest performance on the old dataset Market1501 by \textbf{57.2\%}/\textbf{71.7\%} and the lowest performance on the new dataset CUHK03 by \textbf{41.7\%}/\textbf{42.2\%}, while the performance is reversed when the KCM is employed. This suggests that our FMM achieves better knowledge retention by mapping old knowledge, while the KCM achieves better new knowledge acquisition by continuously matching and weighting knowledge prototypes in new data. Therefore, by combining these two modules, our BiCT achieves the highest performance (reaching \textbf{50.1\%}/\textbf{61.5\%} on mAP/R1) by knowledge balancing.

\par Furthermore, we explored alternative approaches to validate the rationality of our designed transfer architecture. As shown in Table~\ref {tab: ablation_cct_alter}, Full-Conv denotes using a fully convolutional network~\cite{long2015fully}, Full-Linear denotes using a fully connected network, and Full-Trans denotes using a Transformer network~\cite{dosovitskiy2020image} for feature transfer, respectively. It can be seen that our BiCT-Net achieved the highest performance on both mAP/R1. This can be attributed to the fact that our BiCT-Net better approximates the discrepancy between the old and new data distributions through old knowledge mapping and new knowledge capturing, thus achieving better feature transfer and promoting the preservation of old knowledge after transfer compared to alternative approaches.

\noindent\textbf{Effectiveness of BiCD module.}
To evaluate the effectiveness of our BiCD module, we evaluate the performance improvement after introducing BiCD, as shown in Table~\ref{tab: ablation}. It can be seen that our BiCD module improves the average mAP/R1 by \textbf{1.2\%}$\sim$\textbf{1.1\%} in RFL-ReID and L-ReID tasks, respectively. The reason for the above improvement is two-folded. First, the BiCD module promotes the mutual transformation between new and old knowledge through the bidirectional feature alignment, enhancing the consistency of new and old knowledge. Second, the bidirectional distillation of the old relationship maintains the old knowledge in the new feature space, while enabling the new knowledge to preserve more global relationships between different people in the old feature space. Finally, the BiCD module improves the discriminability of the new features while suppressing the domain shift of the updated old features through the bidirectional transfer of new and old knowledge, greatly enhancing its compatibility.

\noindent\textbf{Effectiveness of BiAD module.} 
We further evaluate our proposed BiAD module to verify its anti-forgetting capacity. Table~\ref{tab: ablation} reports the effectiveness of our BiCD module, which achieves average improvements by \textbf{0.7\%}/\textbf{1.2\%} (on mAP/R1 accuracy) in the RFL-ReID task, while maintaining advanced traditional L-ReID performance. This is because our BiAD module promotes the discriminability of old knowledge in the new feature space while enabling the new knowledge to preserve more discriminative information in the old feature space. Therefore, the preserved old knowledge further balance the old and new knowledge and improves the discriminability of both new features and updated old features.

\noindent\textbf{Effectiveness of DFF module.} 
In addition to the above module, the effectiveness of DFF is also verified in Table~\ref{tab: ablation}, where the RFL-ReID performance is further improved by \textbf{1.1\%}/\textbf{0.6\%} on mAP/R1 accuracy without affecting L-ReID performance. Obviously, our DFF module will not contribute to the traditional L-ReID since it is only performed once during the inference phase. In addition, it significantly improves the performance of RFL-ReID due to the proportional fusion within the model parameter space and the updated feature space, achieving knowledge correspondence and eliminating the accumulated forgetting of old knowledge after multi-stage feature transferring.

Besides, to validate the effectiveness of our dynamic fusion strategy, we compared some popular fusion weight $\epsilon^t$ choices, including (a) a fixed parameter with 0.5, (b) a time-increasing parameter with 1-1/t, (c) a time-descending parameter 1/t~\cite{pu2021lifelong, lin2022towards}. As shown in Table~\ref{tab: ablation_fuse}, the results can be observed that our DDF strategy leads all other combination methods commonly used in continual learning on both mAP/R1 accuracy. This can be attributed to the fact that our method employs the same fusion parameters for both the updated ReID model and updated gallery features (\emph{i.e.}, Eq.~\ref{eq:dff_param_fuse} and Eq.~\ref{eq:dff_gallery_ext}), thus achieving better consolidation of knowledge before and after the transfer in a synchronous and compatible manner.

\subsection{Hyper-parameter Study} There are 4 hyper-parameters in our Bi-C\textsuperscript{2}R method, \emph{i.e.} $\mu_1$, $\mu_2$, $\mu_3$, and $\mu_4$. Therefore, we measure the impact of the above parameters, respectively. Fig.~\ref{fig: params} shows the performance of our Bi-C\textsuperscript{2}R method under different hyper-parameter settings. As shown, as $\mu_1$ increases before 100, the new and old knowledge gradually align with each other, while an excessive value will limit the capacity to capture new knowledge, resulting in a decrease in average performance. Meanwhile, $\mu_2$ controls the consistency between the new and old feature spaces, which achieves the highest performance when $\mu_2$=1, avoiding overly strict constraints on the variety between the new and old knowledge. Similarly, we set a reasonable $\mu_3$ (=$7e^{-2}$) and $\mu_4$ (=$5e^{-4}$) to balance the compatibility and the discriminability of the updated features and new features.

\subsection{Visualization}

To intuitively verify the effectiveness of our Bi-C\textsuperscript{2}R, we present more visualization results to intuitively illustrate its effectiveness.

\noindent\textbf{Comparisons on More Training Orders.} More different training orders are helpful to fully verify the robustness under different scenarios. Therefore, following~\cite{pu2023memorizing}, we organize more experiments on different training orders, as shown in Fig.~\ref{fig: orders}. It can be seen that our method shows promising robustness in various scenarios regardless of whether old data can be re-indexed. The above results imply that our Bi-C\textsuperscript{2}R can adaptively capture new and old knowledge in different domains through bidirectional feature transfer, thereby effectively transferring old features and achieving continuous feature updating, demonstrating strong anti-forgetting capacity.

\noindent\textbf{Detailed Performance on Each Training Stage.} Fig.~\ref{fig: stages} illustrates the performance of our method at different training stages. It can be seen that as the number of training stages increases, the performance superiority of our method becomes increasingly obvious, especially in the RFL-ReID task. This is mainly due to the fact that our bidirectional feature updating continuously aligns the upgraded knowledge with the new knowledge, thus avoiding the catastrophic forgetting of old knowledge and the incompatibility between new and old knowledge caused by data domain differences, which finally achieves advanced performance.

\noindent\textbf{T-SNE Results of Gallery and Query Features.} We further employ t-SNE~\cite{van2008visualizing} to intuitively visualize the gallery and query feature extracted by our method compared to the Base method and our previous version. As shown in Fig.~\ref{fig: tsne}, as the old gallery features have not been updated, there is a clear confusion between the old gallery features and the new query features belonging to the same person. Besides, although our previous version (C\textsuperscript{2}R) preliminarily achieved old feature updating, it still suffered from the limited expression of new discriminative knowledge and the accumulative loss of feature compatibility due to the discrepancy between different data domains. In contrast, our Bi-C\textsuperscript{2}R achieves optimal gallery and query feature compatibility via bidirectional feature transfer and dynamic feature fusion, thereby achieving tighter intra-class representation with more discriminative information.

\noindent\textbf{More ReID Results.} Fig.~\ref{fig: reid} visualized the ReID results of our Bi-C\textsuperscript{2}R compared to our Base method. As shown above, the ReID results of Base methods often suffer from similar bodies, appearances, clothing styles and scenarios, due to the catastrophic forgetting of the old discriminative information after multi-stage lifelong learning. However, our method adaptively balances the compatibility between new and old knowledge, so that the discriminative information in the old features can be fully retrieved by the new features after being updated, thereby achieving promising ReID performance in various scenarios.

\section{Conclusion}
In this paper, we focus on a practical and challenging task called Re-indexing Free Lifelong Person Re-identification (RFL-ReID), which prohibits the re-indexing of raw images in the gallery due to data privacy concerns. To this end, we propose a Bidirectional Continual Compatible Representation (Bi-C\textsuperscript{2}R) method, which introduces a Bidirectional Continual Compatible Transfer network to continuously transfer old gallery features to the new feature space. Besides, a Bidirectional Compatible Distillation module and a Bidirectional Anti-forgetting Distillation module are proposed to balance the anti-forgetting of old knowledge with the compatibility to the new model. Extensive experiments on multiple L-ReID benchmarks consists of seven widely-used ReID datasets verify the effectiveness of our method in the RFL-ReID task while maintaining the state-of-the-art performance on the general L-ReID scenario.

\section*{Acknowledgments}
This work was supported by the grants from the National Natural Science Foundation of China (62525201, 62132001, 62432001) and Beijing Natural Science Foundation (L247006).

\ifCLASSOPTIONcaptionsoff
  \newpage
\fi

\bibliographystyle{IEEEtran}
\bibliography{reference}
\newpage
\begin{IEEEbiography}[{\includegraphics[width=1in,height=1.25in,clip,keepaspectratio]{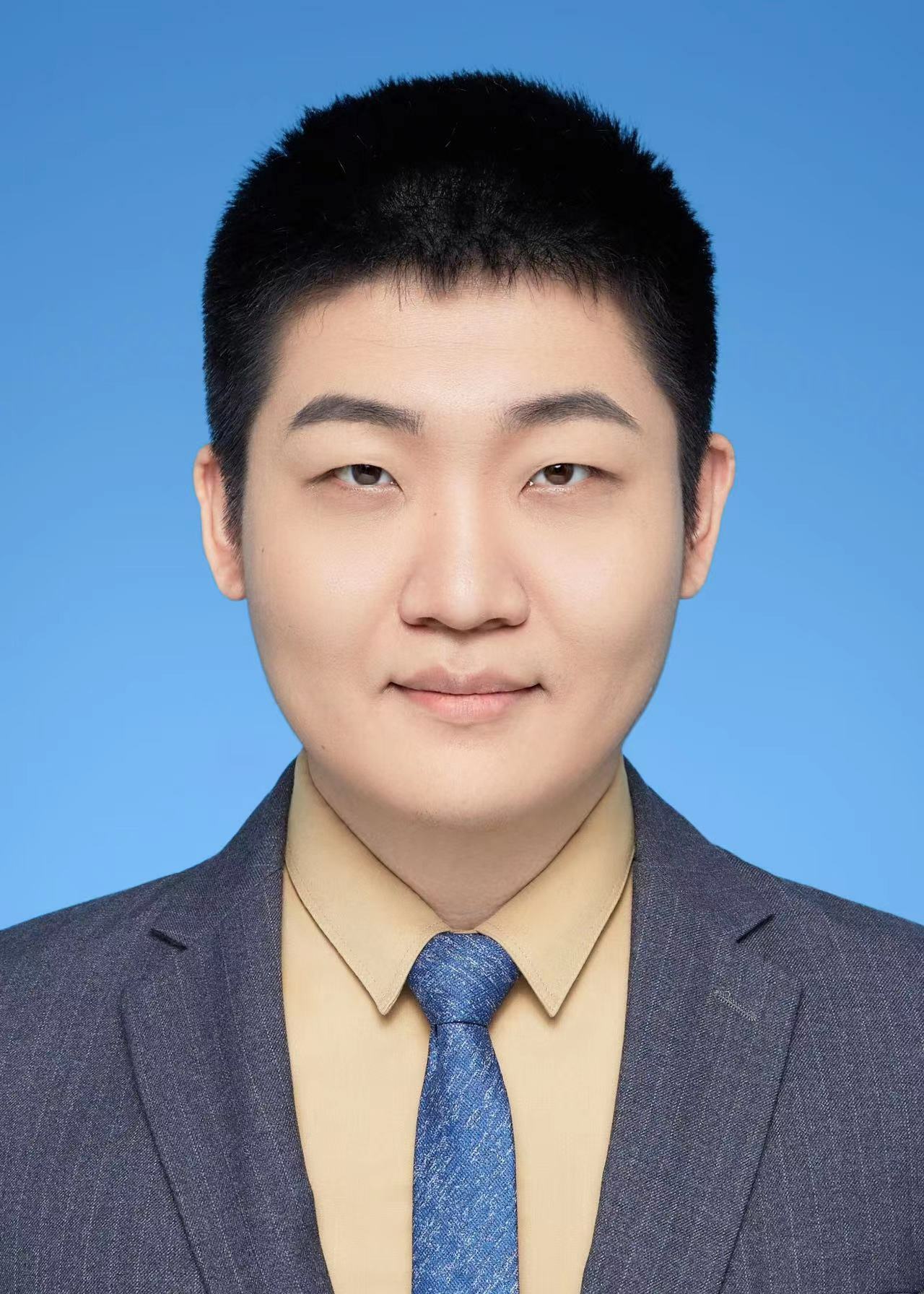}}]{Zhenyu Cui} received the B.S. degree in computer science and technology from China University of Petroleum (East China), Tsingdao, China, in 2018 and the M.S. degree in computer science from University of Chinese Academy of Sciences, Beijing, China, in 2021. He is currently pursuing the Ph.D. degree with the Wangxuan Institute of Computer Technology, Peking University. His current research interests include computer vision and deep learning.
\end{IEEEbiography}
\begin{IEEEbiography}[{\includegraphics[width=1in,height=1.25in,clip,keepaspectratio]{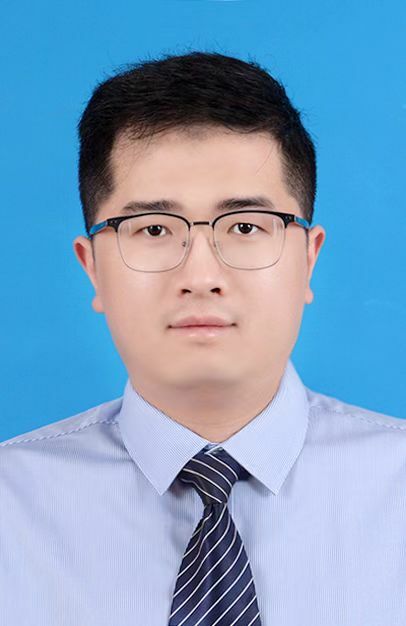}}]{Jiahuan Zhou} received his B.E. (2013) from Tsinghua University, and his Ph.D. degree (2018) in the Department of Electrical Engineering \& Computer Science, Northwestern University. Currently, he is a Tenure-Track Assistant Professor at the Wangxuan Institute of Computer Technology, Peking University. His research interests include computer vision, deep learning, and machine learning. He has authored more than 50 papers in international journals and conferences including Nature Communications, Nature Synthesis, IEEE TPAMI, IJCV, IEEE TIP, IEEE TIFS, CVPR, ICCV, ECCV, ICLR, AAAI, ACM MM, and so on. He serves as an Area Chair for CVPR, ICML, NeurIPS, ICME, and ICPR, an Associate Editor of the Springer Journal of Machine Vision and Applications (MVA), a regular reviewer member for several journals and conferences, e.g., T-PAMI, IJCV, TIP, CVPR, ICCV, ECCV, NeurIPS, ICML, ICLR, AAAI, and so on.
\end{IEEEbiography}
\begin{IEEEbiography}[{\includegraphics[width=1in,height=1.25in,clip,keepaspectratio]{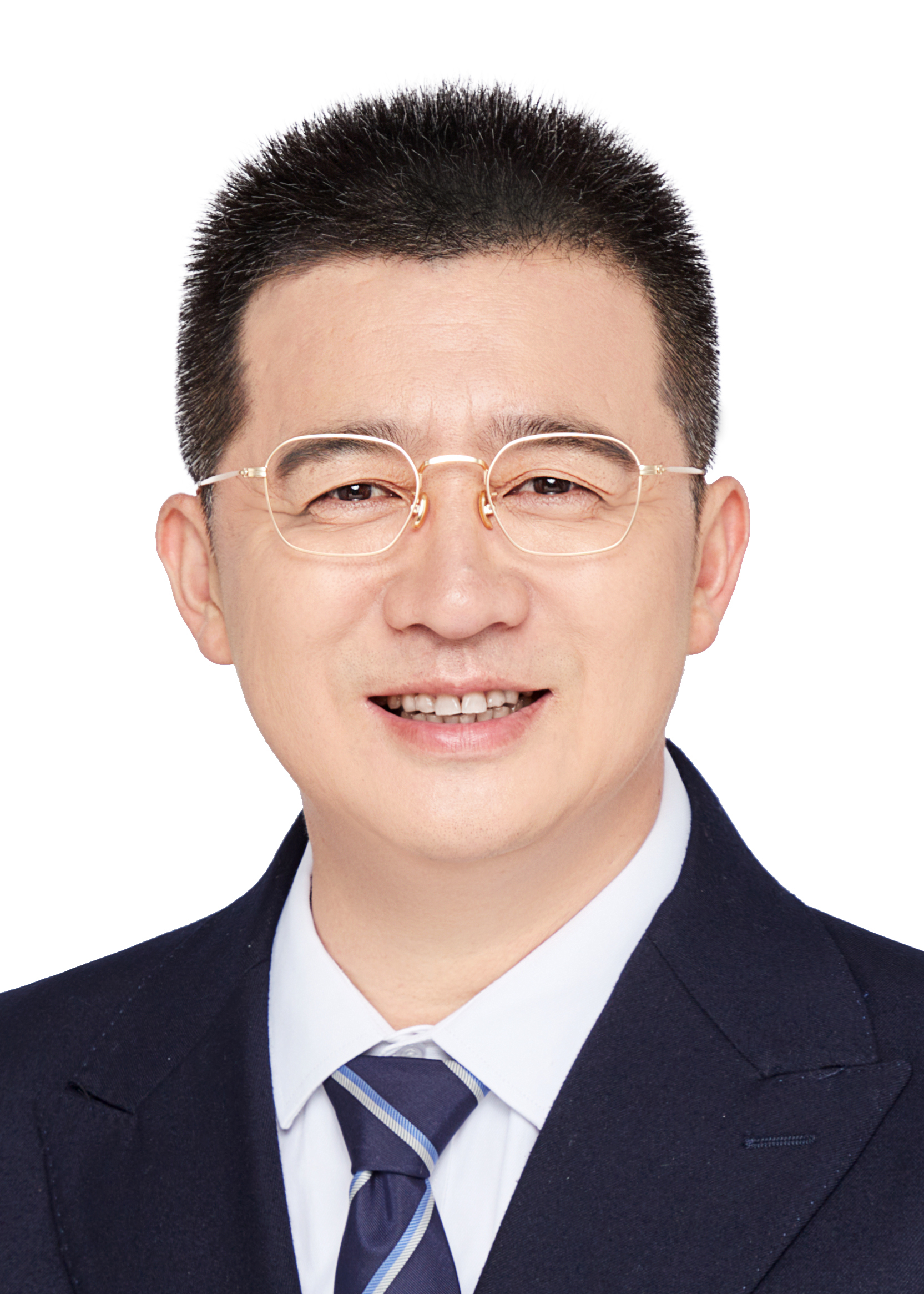}}]{Yuxin Peng}, IEEE/CCF/CAAI/CIE/CSIG Fellow, is the Boya Distinguished Professor at Wangxuan Institute of Computer Technology, Peking University. He was a recipient of the National Science Fund for Distinguished Young Scholars of China in 2019 and its continued funding in 2025. He received the Ph.D. degree in computer application technology from Peking University, Beijing, China, in 2003. His research interests mainly include multimedia analysis, computer vision and artificial intelligence. He has authored over 260 papers, including more than 160 papers in top-tier journals and conference proceedings. He has been granted 39 invention patents. He led his team to win the First Place in the video semantic search evaluation of TRECVID ten times in recent years. He won the First Prize of the Beijing Science and Technology Award in 2016 and the First Prize of the Scientific and Technological Progress Award of the Chinese Institute of Electronics in 2020 as the lead recipient. He was a recipient of the Best Paper award at MMM 2019 and NCIG 2018, and serves as the associate editor of IEEE TMM, TCSVT, etc.
\end{IEEEbiography}
\vfill
\end{document}